\documentclass{article} 
\usepackage{fontenc, inputenc, calc, indentfirst, fancyhdr, graphicx, epstopdf, ifthen, lineno, float, amsmath, setspace, mathpazo, booktabs, xcolor, microtype, tikz, amsthm, subcaption, xcolor, url, hyperref, geometry, newfloat, caption, bm}
\usepackage[square,numbers, sort&compress]{natbib}

\title{Goal-directed Planning and Goal Understanding by Active Inference: Evaluation Through Simulated and Physical Robot Experiments}

\author{Takazumi Matsumoto, Wataru Ohata, Fabien C. Y. Benureau and Jun Tani$^*$}

\date{\small $^1$Okinawa Institute of Science and Technology, 904-045, Japan \\ $^*$Corresponding author: \texttt{jun.tani@oist.jp}}

\begin{document}
\maketitle

\section*{Abstract}
We show that goal-directed action planning and generation in a teleological framework can be formulated using the free energy principle. The proposed model, which is built on a variational recurrent neural network model, is characterized by three essential features. These are that (1) goals can be specified for both static sensory states, e.g., for goal images to be reached and dynamic processes, e.g., for moving around an object, (2) the model can not only generate goal-directed action plans, but can also understand goals by sensory observation, and (3) the model generates future action plans for given goals based on the best estimate of the current state, inferred using past sensory observations. The proposed model is evaluated by conducting experiments on a simulated mobile agent as well as on a real humanoid robot performing object manipulation.

% Keywords
\emph{Keywords: active inference; teleology; goal-directed action planning}

\section{Introduction}
In studying contemporary models of goal-directed actions of biological and artificial agents, it is worthwhile to consider these models from the perspective of a teleological framework. 
Teleology is a philosophical idea that originated in the days of Plato and Aristotle. It holds that phenomena appear not by their causes, but by their end results. 
While teleology is controversial and largely abandoned as a means of explaining physical phenomena, the idea has been extended for modeling action generation. 
A teleological account explains actions by specifying the state of affairs or the event toward which the actions are directed \cite{sehon2007goal, lohrer2016actions}. 
More simply, actions are generated to achieve purposes or goals. 
In psychology, \citeauthor{csibra2003one} \cite{csibra2003one} proposes that a goal-directed action can be explained by a well-formed teleological representation consisting of 3 elements: (1) a goal, (2) actions intended to achieve the goal, and (3) constraints, which are physical conditions that impose limits on possible actions. In a well-formed teleological representation of an action, the action should be seen as an effective means to achieve the goal within the constraint of reality. 
This account predicts that agents who represent goal-directed actions in this way should be able to infer unseen goals or unseen reality constraints, given the two remaining elements. \citeauthor{csibra2003one} \cite{csibra2003one} verified this hypothesis by showing that even year-old infants can perform such inferences in their experiments.

Brain-inspired models for goal-directed action have been developed in various forms.
Most brain-inspired models are based on forward models \cite{kawato1990trajectory, miall1996forward, kawato1999internal} for predicting sensory outcomes of an action to be executed.
For goals given in terms of a preferred sensory state at the distal (terminal) step, optimal action sequences leading to the goal under assumed constraints, such as a minimal torque change criterion, can be obtained inversely using the forward model.
Recently, Friston and colleagues \cite{friston2015active, parr2019generalised} advanced the idea of goal-directed action using a forward model by incorporating a Bayesian perspective.
Goal-directed planning for achieving the preferred sensory states is formulated under the framework of active inference \cite{friston-biol-2011, friston2012active, baltieri2017active} based on the free energy principle \cite{friston2005theory}.

The aforementioned studies on goal-directed action can be expanded in various ways. One such possibility concerns the goal representation.
Although goals in the aforementioned models are represented by a particular sensory state at each time step or at the distal step, some classes of on-going processes can also be goals for generating actions.
For example, one can say "I got up early this morning to run."
In this case, the goal is not a particular destination, but the process of running.
In the teleological framework, goals or the purpose of actions could be states, events, or processes which could be either specific or abstract and conceptual.
What sorts of models could incorporate such diverse forms of goal representations? 
Another possibility for exploration is to incorporate the capability for inferring unseen goals or unseen reality constraints provided with the remaining two elements among actions, constraints and goals, as described by \citeauthor{csibra2003one} \cite{csibra2003one}.

For the purpose of examining these possibilities, the current study proposes a novel model by extending our prior goal-directed planning model, GLean \cite{Matsumoto_2020}.
GLean was developed following the free energy principle \cite{friston2005theory} and it operates in the continuous domain using PV-RNN, a variational recurrent neural network model \cite{ahmadi2019novel}.
The newly proposed model, T-GLean, has three new key features compared to GLean.
First, goals can be specified either by temporal processes (such as the earlier example of "going out for a run") or static sensory states, for example, "going to the corner store."
Second, the model can infer goals by sensory observation, as well as by generating goal-directed action plans. This allows agents using the T-GLean model to anticipate goals and future actions by observing the actions of another agent.
Third, the model generates future action plans for given goals based on the best estimate of the current hidden state using the past sensory observation. While this feature is not necessarily novel, given that functions of sensory reflection or postdiction \cite{shimojo2014postdiction} have been examined using deterministic generative RNN models \cite{tani2003learning} and Bayesian probabilistic generative models \cite{parr2019generalised}, this feature is added to the model so that robots utilizing T-GLean can generate goal-directed plans online while actions are being executed, whereas GLean can only generate plans offline.

Our proposed model is evaluated in Section~\ref{sec:experiments} by conducting two experiments using two robot setups. 
The first experiment uses a simple simulated mobile robot that can navigate an environment with an obstacle, and the second experiment uses a physical humanoid robot with a larger degree of freedom that tracks and manipulates objects.
In both experiments, the robots are trained to achieve two different types of goals.
For the mobile robot, one type of goal is to reach a specified position while avoiding collisions with obstacles, and the other type of goal is to move around a specific object repeatedly.
For the humanoid robot, one type of goal is to grasp an object and then to place it at a specified position and the other type of goal is to lift an object up and down repeatedly.
In both experiments, the trained robots are evaluated in generating goal-directed actions for specified goals, as well as to infer corresponding unseen goals for given sensory sequences. We also touch on generating action plans based on the rationality principle \cite{csibra2003one}.
The following section describes related studies in more detail so that readers can understand easily how the proposed model has been developed by extending and modifying those prior proposed models.

\section{Related studies}
\begin{figure}[H] % Fig. 1
\subcaptionbox{\label{fig:related-models-fwd}}{\includegraphics[width=0.33\linewidth]{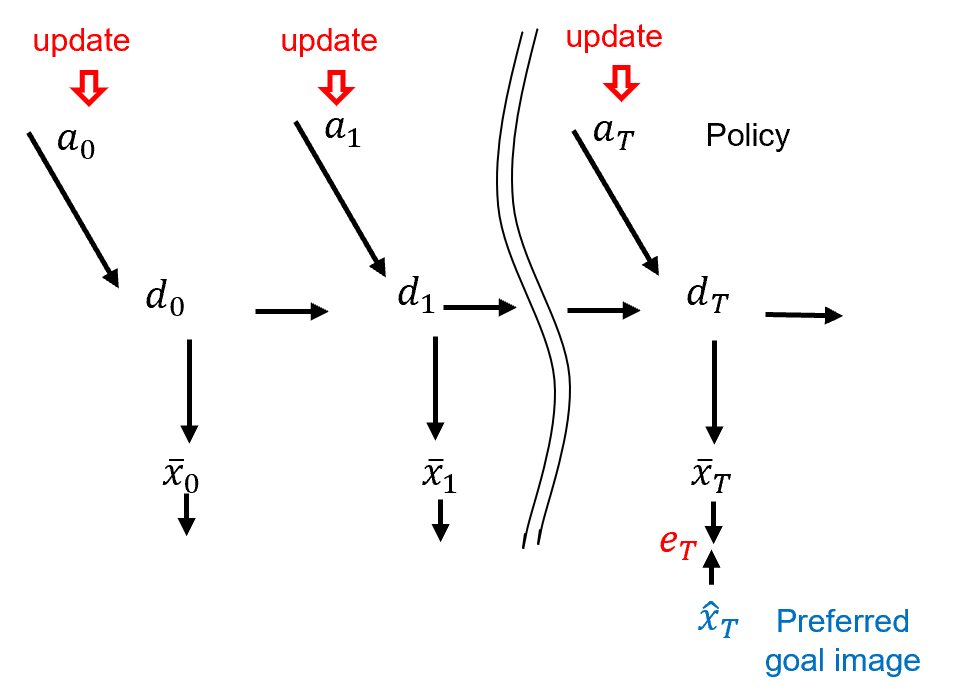}}
\subcaptionbox{\label{fig:related-models-aif}}{\includegraphics[width=0.33\linewidth]{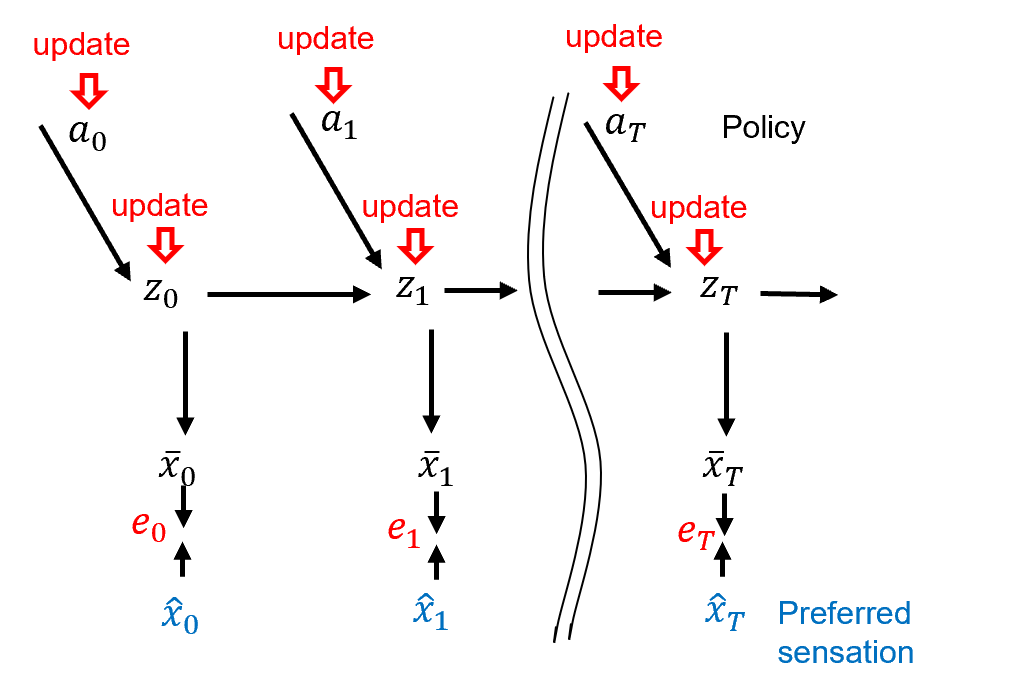}}
\subcaptionbox{\label{fig:related-models-glean}}{\includegraphics[width=0.33\linewidth]{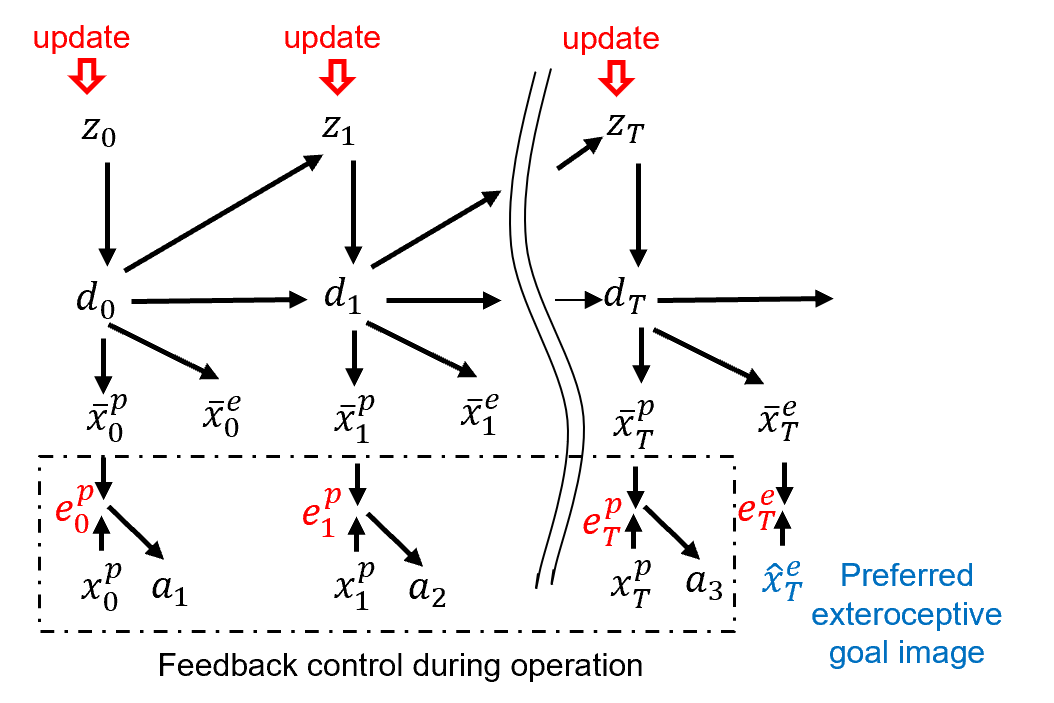}}
\caption{Prior models for generating goal-directed behaviors. (\textbf{a}) Forward model using latent variables, (\textbf{b}) active inference model using probabilistic latent variables, and (\textbf{c}) GLean as an extension of active inference.}
\label{fig:related_models}
\end{figure}

First we look in detail at how a goal-directed action can be generated using the forward model.
\citeauthor{kawato1990trajectory} \cite{kawato1990trajectory} proposed that multiple future steps of proprioception (joint angles) as well as the distal position in task coordinate space of an arm robot can be predicted, given the motor torque at each time step, by training the forward dynamics model and the kinematic model that are implemented in a feed-forward network cascaded in time.
After training, optimal motor commands to achieve the desired positions at the distal step following the minimal torque change criterion can be computed by using the prediction error information generated in the distal step.
In order to deal with the hidden state problem encountered by real robots navigating with limited sensors, \citeauthor{tani1996model} \cite{tani1996model} extended this model by implementing the forward dynamics model in a recurrent neural network (RNN) with deterministic latent variables.
It was shown that a physical mobile robot can generate optimal, goal-directed navigation plans with the minimum travel distance criterion in a hidden state environment using that proposed model.
Figure~\ref{fig:related-models-fwd} depicts the scheme of goal-directed planning using a forward dynamics model implemented in an RNN.
In this graphical representation, $a_t$, $d_t$, $z_t$ and $\bar{x}_t$ denote the action, deterministic latent variable, probabilistic latent variable, and sensory prediction at time step $t$, respectively. $\hat{x}_{T}$ is the sensory goal image at the distal step $T$.
Within the scheme of the forward model and active inference, the policy is a sequence of actions that is optimized such that the error between the sensory goal image and the preferred image at $T$ can be minimized (the minimization criterion such as the travel distance is omitted for brevity.)

Friston and colleagues \cite{friston2015active, parr2019generalised} formulated goal-directed actions by extending the framework of active inference \cite{friston-biol-2011, friston2012active, baltieri2017active} based on the free energy principle \cite{friston2005theory}.
The major advantage of this approach, compared to the aforementioned conventional forward model, is that the model can cope with uncertainty that originates from the stochastic nature of the environment by incorporating a Bayesian framework.
Originally, the free energy principle was developed as a theory for perception under the framework of predictive coding, wherein probabilistic distributions of past latent variables in generative models are inferred for the observed sensation by minimizing the free energy.
Later, active inference was developed as a framework for action generation wherein action policies as well as the probabilistic distribution of future latent variables in generative models are inferred by minimizing the so-called expected free energy.
Minimizing the expected free energy $G_{aif}$ for the future maximizes both the epistemic value shown in the first and the second terms, and extrinsic value in the third term in Equation \ref{eq:G_}.

\begin{equation} \label{eq:G_}
    G_{aif} = \sum_{t>{t_c}}^T -E_{q({z_t},{x_t}|\pi)} \left[\; \; \; \underbrace{\log p(z_t | x_t) - \log q(z_t|\pi)}_{\text{epistemic value}} \; \; + \underbrace{\log P(x_t)}_{\text{extrinsic value}}\right] 
\end{equation}

where $\pi$ is an optimal policy or a sequence of actions, $q()$ is the posterior predictive distribution, and $t_c$ is the current time step.
The extrinsic value represents how much the sensory outcomes expected in the generated future plan are consistent with the preferred outcomes. 
The epistemic value, on the other hand, represents the expected information gain with predicted outcomes. This means that this value indicates the expected decrease of uncertainty with regard to the hidden state provided by the sensory observation.
In summary, an optimal action policy for achieving the preferred sensory outcomes and the reduction of uncertainty of the hidden states can be generated by minimizing the expected free energy. 
Figure \ref{fig:related-models-aif} depicts this framework wherein the probabilistic latent variables $z_{0:T}$, as well as an action policy, $a_{0:T}$ are inferred by minimizing the expected free energy. 

Although Friston and colleagues \cite{friston2015active, parr2019generalised} implemented this framework mostly in discrete space, \citeauthor{hafner2019dream} \cite{hafner2019dream} proposed an analogous model implemented in continuous state problems.

\citeauthor{Matsumoto_2020} \cite{Matsumoto_2020} proposed another type of goal-directed action plan generation model, GLean, which is based on the free energy principle using a variational RNN.
The main difference of this scheme from those proposed by Friston and colleagues \cite{friston2015active, parr2019generalised} and by \citeauthor{hafner2019dream} \cite{hafner2019dream} is that an optimal goal-directed plan is obtained by inferring the lower dimensional probabilistic latent space, instead of the higher dimensional action space.
The graphical model of the scheme is depicted in Figure \ref{fig:related-models-glean}, wherein the proprioception $\bar{x}^p_{t}$ and the exteroception $\bar{x}^e_{t}$ at each time step $t$ are predicted by the learned generative model using the probabilistic latent variable $z_t$ and deterministic latent variable $d_t$.
For a given preferred goal represented as an exteroceptive state at the distal step $T$ as ${\hat{x}_T}$, a proprioceptive-exteroceptive sequence expected to reach this goal state is searched by inferring an optimal sequence of the posterior distribution of the latent variables $q(z_{1:T})$ by means of minimizing the expected free energy $G_{gl}$ as shown in Equation~\ref{eq:G_gl}.

\begin{equation} \label{eq:G_gl}
    G_{gl} = \underbrace{-E_{q({z}_T | {\hat{x}_T)}} \big[ p({x}_T | {d}_{T}) \big]}_{\text{goal error}}
    + \underbrace{\Big(\sum_{t=t_c}^T D_{KL}\big[ q({z}_t | \hat{x}_T) || p({z}_t | {d}_{t-1}) \big] \Big)}_{\text{complexity}}
\end{equation}
Here, the expected free energy is represented by a sum of two terms, namely the goal error at the distal step, shown in the first term, and the complexity summed over all future steps, shown in the second term.
Complexity is the divergence between the posterior predictive distribution and the prior distribution of the probabilistic latent variables, and is described in more detail later in Section~\ref{sec:model}.
By minimizing this form of free energy, plans reaching the preferred goal states, but following well-habituated trajectories learned during learning, can be generated.
We note that this model does not infer the policy directly. Instead, the motor command or action $a$ at each time step is computed by means of the proprioceptive error feedback by sending the prediction of the next step proprioception as the target to the motor controller when the agent generates movement.
%%% End of related study section

\section{The Proposed Model} \label{sec:model}
\begin{figure}[H]
\centering
\includegraphics[width=0.33\linewidth]{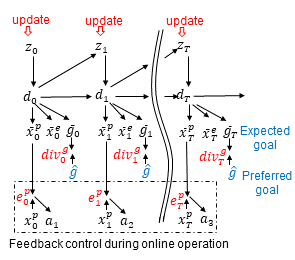}
\caption{Overview of the newly proposed model.}
\label{fig:new-model-simple}
\end{figure}

Our newly proposed model, T-GLean, is an extension of GLean \cite{Matsumoto_2020} with a novel goal representation scheme.
Each goal is represented by its category, e.g., reaching or cycling, associated with optional parameters,  e.g., for position or speed.
The network is designed to output the expected goal at each time step based on learning, as shown in Figure~\ref{fig:new-model-simple}.
When the current preferred goal is given at each time step, the posterior predictive distribution is updated in the direction of minimizing the divergence between the preferred goal and the expected goal at each time step.
This generates the expected exteroceptive and proprioceptive trajectory leading to the goal.
In addition, the network can infer unseen goals from the observation of on-going sensory inputs.
For a given exteroceptive trajectory, the posterior distribution is updated in the direction of minimizing the divergence between the given exteroceptive trajectory and its reconstructed trajectory.
This generates the expected goal in the outputs at each time step.
This model utilizes the PV-RNN architecture, which leverages the idea of multiple timescale RNN \cite{yamashita2008emergence,Matsumoto_2020} with the expectation of development of a functional hierarchy through learning.
The network is operated in three modes, each of which minimizes free energy as its loss function.
In learning mode, predictive models are learned using provided training sequences by minimizing the evidence free energy.
In online action generation mode, a future goal-directed plan is generated by minimizing the expected free energy while the past sensory experience is reflected by minimizing the evidence free energy. As this occurs in real time, the past sensory experience is constantly updated online. 
In the goal inference mode, the expected goal is inferred using the observed exteroceptive sequence by minimizing the evidence free energy.
The following sub-sections describe the model in more detail.

\subsection{Model architecture}

\begin{figure}[H]
\centering
\includegraphics[width=0.8\linewidth]{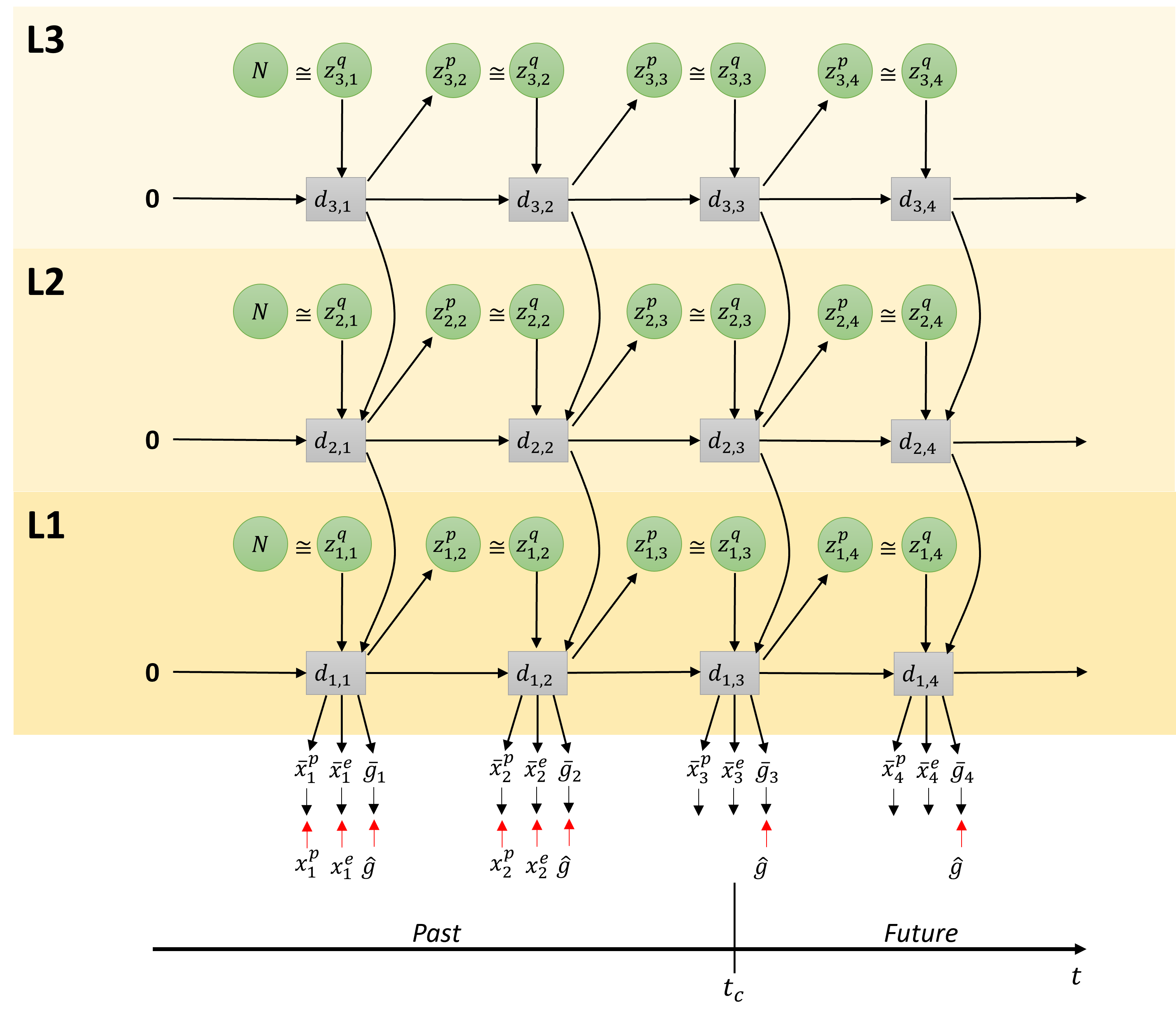}
\caption{Graphical description of the employed architecture.}
\label{fig:architecture}
\end{figure}

Figure \ref{fig:architecture} depicts the employed architecture as a graphical model consisting of three layers of PV-RNN. It is similar to the architecture employed in \cite{Matsumoto_2020}, with the modification introduced in \cite{ohata2020} that has the top-down connection from higher layers to lower layers $d_{l+1} \rightarrow d_l$ in the same time step, rather than from the previous time step.
The graphical model shows the RNN unrolled into to the past as well as to the future, with the current time step at $t_c$.
Each layer indexed by $l$ contains the stochastic variable $z$, from which $z_{l,t}^p$ is sampled from the prior distribution, and $z_{l,t}^q$ is the posterior distribution, as well as the deterministic latent variable $d_{l,t}$, for each time step $t$. Note that deterministic variables take as input the deterministic variable from the layer above except on the top layer.
The output of the bottom layer (L1) is split into proprioception $\bar{x}^p_t$, exteroception $\bar{x}^e_t$, expected goal $\bar{g}_t$, and distal probability $\bar{s}_t$. Unless noted otherwise, the preferred goal $\hat{g}$ is given at all time steps as a target. Where available, the observed proprioception $x^p_t$ and exteroception $x^e_t$ are used as targets for error minimization.
The distal probability at time step $\bar{s}_t$ (omitted from the diagram for brevity) is the probability of achieving the expected goal at $t$. 
When the distal probability at a particular time step becomes the highest among those at all other time steps and it exceeds a predefined threshold value, it is assumed that the preferred goal is achieved in this time step.
During training, $\hat{s}$ is a one-hot vector in the time dimension with a single peak at the time step the goal is actually achieved.
The output $(\bar{\bm{x}}, \bar{g})$ and targets $(\bm{x}, \hat{g})$ are softmax encoded; however,  for brevity, the output layer that decodes the network output into raw output for the agent is not shown in this figure.
In subsequent subsections, we will describe in more detail the aforementioned modes on which the network can operate. We do not make a complete derivation of PV-RNN in this paper, but focus on key aspects of this model.

\subsection{Learning}
Based on the graph connectivity shown in Figure~\ref{fig:architecture}, the forward computation of PV-RNN is given in Equation~\ref{eq:cell}. The internal state of each PV-RNN cell $h_{l,t}$ at time step $t$ and level $l$ is computed as a sum of the connectivity weight multiplication of $z_{l,t}$, $h_{l,t-1}$, and $h_{l+1,t}$ (if $l$ is not the top layer). The connectivity weight matrix $\bm{W}$ is indexed from layer to layer and from unit to unit. For brevity, bias terms have been omitted.

\begin{equation} \label{eq:cell} % (3)
\begin{aligned}
    h_{l, t} &= \left(1 - \frac{1}{\tau^l}\right)h_{l,t-1} + \frac{1}{\tau^l} \left( \bm{W}^{l,l}_{d,d} d_{l,t-1} + \bm{W}^{l,l}_{z,d} z_{l,t} + \bm{W}^{l+1,l}_{d,d} d_{l+1, t} \right) ,\\
    d_{l, t} &= \text{tanh}(h_{l,t}) ,\\
    d_{l, 0} &= 0 .\\
\end{aligned}
\end{equation}
Where $\tau^l$ is the MTRNN time constant of layer $l$. 

The stochastic variable $z$ follows a Gaussian distribution. Each sample of the prior distribution $z_t^p$ for layer $l$ is computed as shown in in Equation~\ref{eq:p}.

\begin{equation} \label{eq:p} % (4)
\begin{aligned}
    \mu^p_{l,t} &=
    \begin{cases}
        0, & \text{if } t=1 \\
        \text{tanh}(\bm{W}^{l,l}_{d,z^{\mu^p}}d_{l,t-1}), & \text{otherwise}
    \end{cases}
    \\
    \sigma^p_{l,t} &=
    \begin{cases}
        1, & \text{if } t=1 \\
        \exp(\bm{W}^{l,l}_{d,z^{\sigma^p}}d_{l,t-1}), & \text{otherwise}
    \end{cases}
    \\
    z^p_{l,t} &= \mu_{l,t}^p + \sigma_{l,t}^p \odot \epsilon .
\end{aligned}
\end{equation}

Where $\epsilon$ is a random noise sample such that $\epsilon \sim \mathcal{N}(0, I)$.
Samples of the posterior distribution $z_t^q$ are computed as shown in Equation~\ref{eq:q}. The $\bm{A}$ variable is a vector for each $z$ unit and sequence. For brevity, here we assume we have a single sequence.
If an output sequence is generated using the posterior adapted during training, the corresponding training sequence should be regenerated.
During goal inference and plan generation, the $\bm{A}$ variables are inferred by the error regression process, as will be described later.
\begin{equation} \label{eq:q} % (5)
\begin{aligned}
    \mu^q_{l,t} &= \text{tanh}(\bm{A}^\mu_{l,t}) ,\\
    \sigma^q_{l,t} &= \exp(\bm{A}^\sigma_{l,t}) ,\\
    z^q_{l,t} &= \mu_{l,t}^q + \sigma_{l,t}^q \odot \epsilon .\\
\end{aligned}
\end{equation}
To compute the output at time step $t$, there are three steps. First, the network output $o$ is computed from $d_{1, t}$, as in Equation~\ref{eq:o}. This uses the output from layer 1 and treats the output layer as layer 0.
\begin{equation} \label{eq:o} % (6)
    o_t = \bm{W}^{1,0}_{d,o} d_{1,t} .
\end{equation}
We then compute the predicted probability distribution output $\bar{x}$.
For this purpose, we use a softmax function to represent the probability distribution of the $i$-th dimension of the output as in Equation~\ref{eq:xsm}.
\begin{equation} \label{eq:xsm} % (7)
    \bar{x}^{i,j}_t = \frac{\exp(o^{i,j}_t)}{\sum_j \exp(o^{i,j}_t)} .
\end{equation}
where $\bar{x}^{i,j}_t$ is the predicted probability that the $j$-th softmax element of the $i$-th output is on.

\begin{figure}[H] % Fig. 4
\centering
\includegraphics[width=0.8\linewidth]{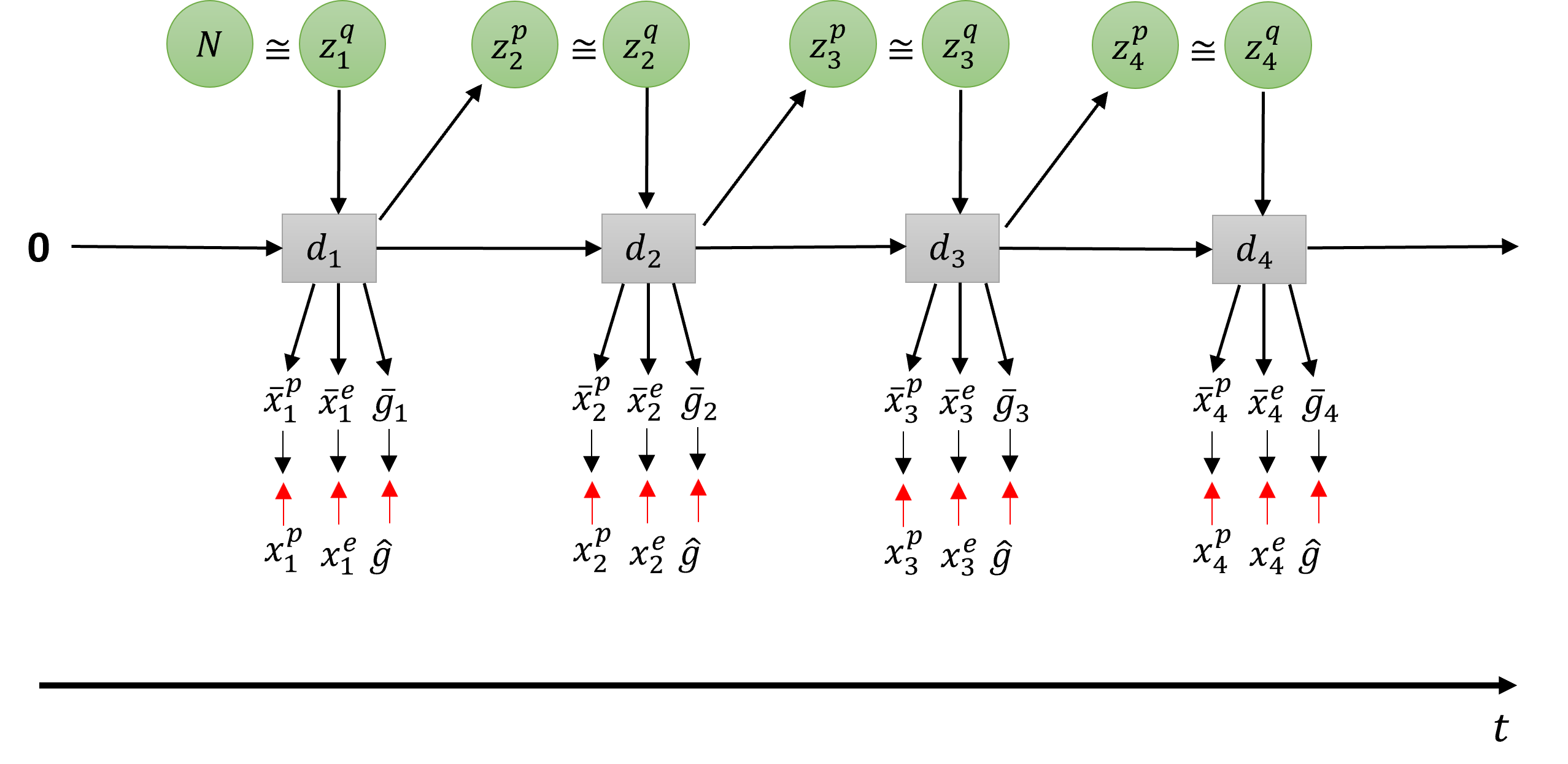}
\caption{Network during training.}
\label{fig:arch1ltrain}
\end{figure}

For explaining the learning scheme following the principle of free energy minimization, we first describe the model in terms of free energy. 
For brevity, we will assume there is a single layer only. This is shown graphically in Figure~\ref{fig:arch1ltrain}.
During learning, the evidence free energy shown in Equation~\ref{eq:train_fe} is minimized by iteratively updating the posterior of $z$ as well as the RNN learning parameters $W$ at each time step for all training sequences.
\begin{equation} \label{eq:train_fe} % (8)
    F(\bm{x}, \hat{g}, z) = \sum^T_{t=1} \Big( w \cdot \underbrace{D_{KL}\big[ q(z_t | \bm{x}_{t:T}, \hat{g}_{t:T}) || p(z_t | d_{t-1}) \big]}_{\text{complexity}} - \underbrace{E_{q(z_t | \bm{x}_{t:T}, \hat{g}_{t:T})} \big[ \log P(\bm{x}_t, g_t | d_t)}_{\text{accuracy}} \big] \Big).
\end{equation}

Where $\bm{x}$ and $\hat{g}$ are the observed sensory states and preferred goal, respectively, while $d$ and $z$ are the deterministic and probabilistic latent states, respectively. 
Free energy in this work is modified by inclusion of the meta-prior $w$, which weights the complexity term. $w$ is a hyperparameter that affects the degree of regularization, similar to $\beta$ in Variational Autoencoders \cite{Kingma2014}. We also note that since we are dealing with sequences of actions, the free energy is a summation over all time steps in the sequence. 

The first term, complexity, is computed as the Kullback–Leibler (KL) divergence between the prior and approximate posterior distributions. This can be expressed as follows:
\begin{equation} \label{eq:train_kld1} % (9)
    D_{KL}\big[ q(z_t | \bm{x}_{t:T}, \hat{g}_{t:T}) || p(z_t | d_{t-1}) \big] = \int q(z_t | \bm{x}_{t:T}, \hat{g}_{t:T}) \log \frac{q(z_t | \bm{x}_{t:T}, \hat{g}_{t:T})} {p(z_t | d_{t-1})} dz_t
\end{equation}
Since we have $\mu$ and $\sigma$ for both prior $p$ and posterior $q$ distributions, they can be expressed as follows:
\begin{equation} \label{eq:train_pq} % (10)
\begin{aligned}
    p(z_t | d_{t-1})                     &= \frac{1}{\sqrt{2\pi(\sigma^p_t)^2}} \exp \Bigg[ - \frac{1}{2} \Bigg( \frac{z_t-\mu^p_t}{\sigma^p_t} \Bigg)^2 \Bigg], \\
    q(z_t | \bm{x}_{t:T}, \hat{g}_{t:T}) &= \frac{1}{\sqrt{2\pi(\sigma^q_t)^2}} \exp \Bigg[ - \frac{1}{2} \Bigg( \frac{z_t-\mu^q_t}{\sigma^q_t} \Bigg)^2 \Bigg]. \\
\end{aligned}
\end{equation}
Thus, continuing from Equation~\ref{eq:train_kld1}, complexity can be computed as:
\begin{equation} \label{eq:train_kld2} % (11)
    \int q(z_t | \bm{x}_{t:T}, \hat{g}_{t:T}) \log \frac{q(z_t | \bm{x}_{t:T}, \hat{g}_{t:T})} {p(z_t | d_{t-1})} dz_t = \log \frac{\sigma^p_t}{\sigma^q_t} + \frac{(\mu^q_t - \mu^p_t)^2 + (\sigma^q_t)^2}{2(\sigma^p_t)^2} - \frac{1}{2}
\end{equation}
For brevity, a case of a single z-unit with a $(\mu, \sigma)$ pair is shown here.
In practice we can have multiple z-units, each with independent $(\mu, \sigma)$, and in that case the RHS is a summation over all $(\mu$, $\sigma$), as will be shown later in the experimental section.
The second term of Equation~\ref{eq:train_fe}, accuracy, can be computed using the probability distribution estimated in the softmax outputs.
During learning, Equation~\ref{eq:train_fe} is used as the loss function with the Adam optimizer, as noted Section~\ref{sec:experiments}.

% Online goal-directed action plan generation
\subsection{Online goal-directed action plan generation}
\begin{figure}[H] % Fig.5
\centering
\includegraphics[width=0.8\linewidth]{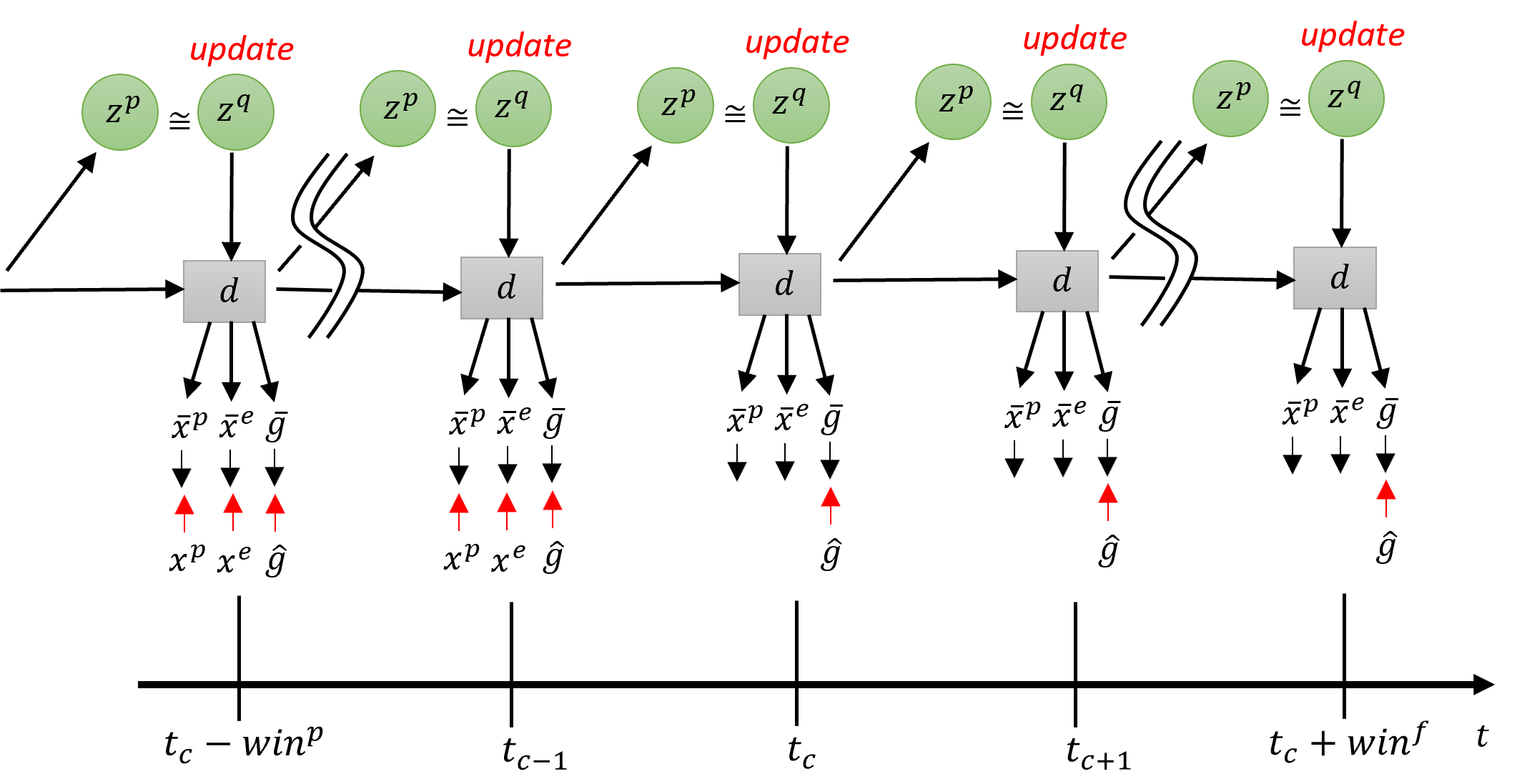}
\caption{The network during planning with a planning window covering from $t_c - win^p$ to $t_c + win^f$ where $t_c$ is current time step.}
\label{fig:planwindow}
\end{figure}

A key difference between our newly proposed model, T-GLean, and our previous model, GLean, is the idea that the goal expectation $\bar{g}_t$ is generated at every time step instead of expecting the goal sensory state at the distal time step.
Intuitively, this means that at every time step, the agent expects a goal state that the on-going action sequence will achieve. 
An advantage of this model scheme is that goals can be represented not only by distal sensory states to be achieved, but also by continuously changing sensory sequences.
Another key feature of T-GLean is that the model generates plans online, in real-time, while the agent is acting on the environment, whereas our prior study using GLean showed only an offline plan generation scheme.
In T-GLean the network maintains the observed sensory sequence in a past window while allocating a future window for the future time steps.
In the past window, evidence free energy is minimized online by updating the posterior at each time step in order to situate all the latent variables to the observed sensory sequence.
In the future window, the error between the preferred goal and the expected goal output is minimized for all steps by updating the posterior predictive distribution in the window iteratively, of which computation can be performed by minimizing the expected free energy.
The future plan for achieving the preferred goals can be generated once the evidence free energy in the past window is minimized, i.e., the latent variables are well situated to the past sensory observation. 
This scheme is referred to as online error regression \cite{tani2003learning} and analogous models can be seen also in \cite{butz2019learning,parr2019generalised}.
The scheme is shown graphically in Figure~\ref{fig:planwindow}. 

Within the planning window, there is the past window of length $win^p$ and the future window of length $win^f$. In the current implementation, the length of the planning window is fixed, while the past window is allowed to grow up to half the length of the planning window. The current time step $t_c$ is one step ahead of the end of the past window. At the next sensorimotor time step, sensory information at $t_c$ becomes part of the past window, and $t_c$ moves forward one step, shrinking the future window. Once the past window is filled, the entire planning window slides one step to the right, discarding the oldest entry in the past window.
During online planning, the network minimizes the plan free energy $F_{\text{plan}}$ by optimizing the posterior $z^q$ at all time steps within the past and future windows. 
The plan free energy consists of the sum of the evidence free energy $F_e$ within the past window and the expected free energy $G$ within the future window. This is expressed in Equation~\ref{eq:plan_esfe}. %Note: This is called estimated free energy in the code
\begin{equation} \label{eq:plan_esfe} % (12)
\begin{aligned}
    F_e(\bm{x}, \hat{g}, z) &= \sum^{t=t_c}_{t=t_c-win^p} \Big( w \cdot D_{KL}\big[ q(z_t | \bm{x}_{t:t_c}, \hat{g}_{t:t_c}) || p(z_t | d_{t-1}) \big] - E_{q(z_t | \bm{x}_{t:t_c}, \hat{g}_{t:t_c})} \big[ \log P(\bm{x}_t, g_t | d_t) \big] \Big), \\
    G(\hat{g}, z) &= \sum^{t=t_c+win^f}_{t=t_c} \Big( w \cdot D_{KL}\big[ q(z_t | \hat{g}_{t:t_c+win^f}) || p(z_t | d_{t-1}) \big] - E_{q(z_t | \hat{g}_{t:t_c+win^f})} \big[ \log P(\bar{g}_t | d_t) \big] \Big), \\
    F_{\text{plan}} &= F_e + G. \\
\end{aligned}
\end{equation}
%
% Goal Inference
\subsection{Goal inference}
\begin{figure}[H] % Fig.6
\centering
\includegraphics[width=0.8\linewidth]{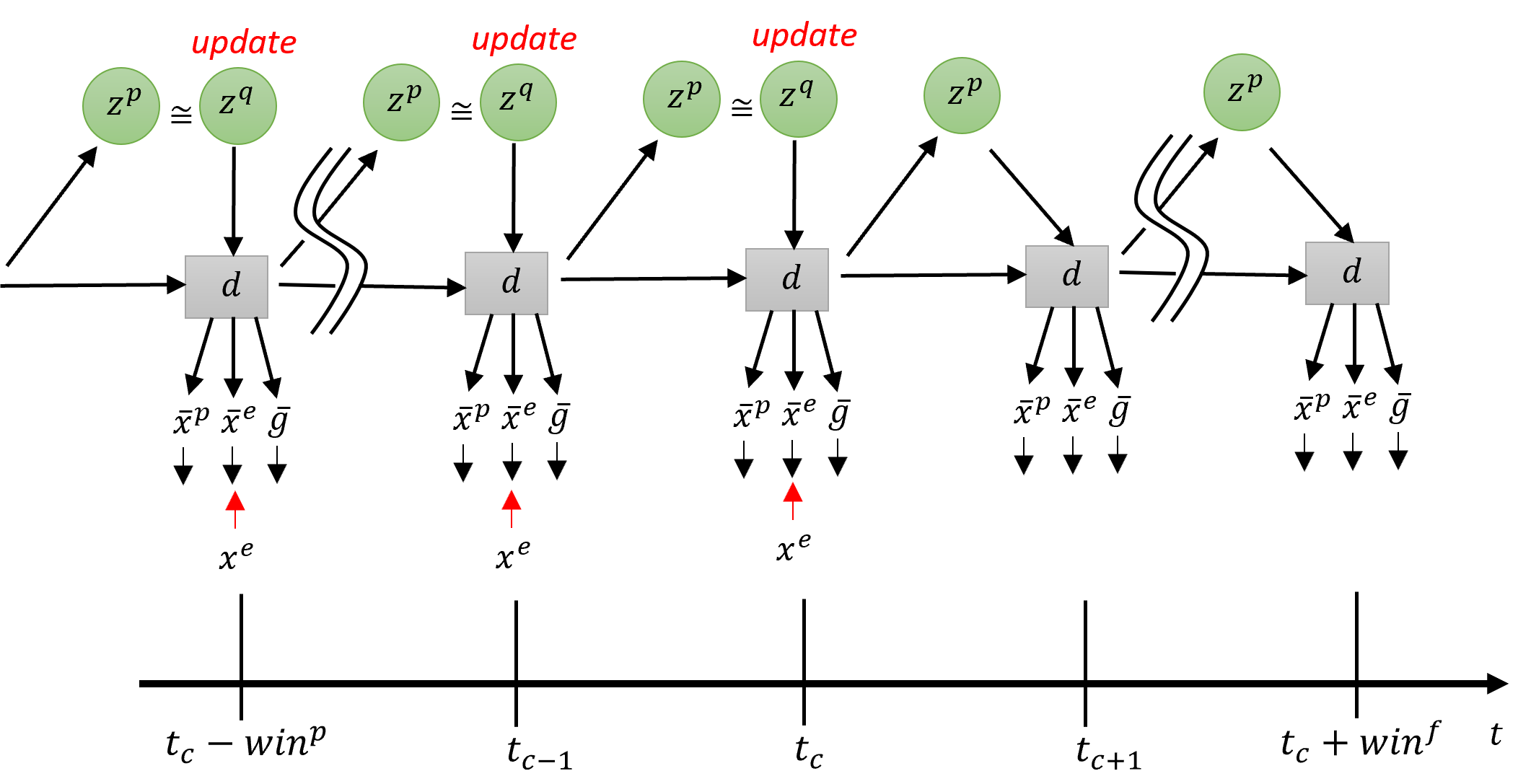}
\caption{Network during goal inference.}
\label{fig:arch1lgoal}
\end{figure}

Finally, before closing the current model section, we describe how future goals can be inferred from observed sensory sequences.
Figure~\ref{fig:arch1lgoal} shows a graphical model accounting for the mechanism of the goal inference with observation of the exteroception, but without actually generating actions.
By observing the exteroception sequence from time step $t_c - win^p$ to the current time step $t_c$, the posterior $z^q$ at each step in the past window is optimized for minimizing the error between the observed and reconstructed exteroceptions.
This results in inference of the expected goal $\bar{g}_t$ for every time step $t$, both in the past and future.
The network predicts simultaneously both the exteroception and proprioception for future steps leading to the inferred goal.
We note that this scheme could be applied to the problem of inferring goals of other agents through observation of their movements, provided that the coordinate transformation between the allocentric view and the egocentric view can be made.
For simplicity, the current study does not delve into this view coordinate transformation problem.
Equation~\ref{eq:goalinf_fe} shows the modified evidence free energy used for goal inference, where only the observation of exteroception is used.
\begin{equation} \label{eq:goalinf_fe} % (13)
    F_g(x^e, z) = \sum^{t=t_c}_{t=1} \Big( w \cdot D_{KL}\big[ q(z_t | x^e_{t:t_c}) || p(z_t | d_{t-1}) \big] - E_{q(z_t | x^e_{t:t_c})} \big[ \log P(x^e_t | d_t) \big] \Big).
\end{equation}

%%%%%%%%%%%%%%%%%%%%%%%%%%%%%%%%%%%%%%%%%%
% Experiments
%%%%%%%%%%%%%%%%%%%%%%%%%%%%%%%%%%%%%%%%%%
\section{Experiments} \label{sec:experiments}
In order to test our proposed model, we conducted two experiments, one using a simulated agent and the other using a physical robot. 
In Experiment 1 (Section~\ref{sec:exp1}), we used a simulated mobile agent in 2D space in order to examine the model's capacity in generating goal-directed actions and in understanding goals from sensory observation. 
Experiment 2 (Section~\ref{sec:exp2}) was carried out to test the model's scalability in the real world setting by using a humanoid robot with higher degrees of freedom.
T-GLean is implemented using LibPvrnn, a custom C++ library implementing PV-RNN that is currently under development. It is designed to be lightweight and to operate in real-time so interaction between agent and experimenter is possible. A pre-release version is \href{https://github.com/oist-cnru/T-GLean}{available} under an open source license with instructions on reproducing the simulated agent experiments.
\subsection{Experiment 1: simulated mobile agent in a 2D space} \label{sec:exp1}
We conducted a set of experiments using a simulated mobile agent that can generate goal-directed action plans based on supervised learning of sensorimotor experiences in order to evaluate the performance of T-GLean in the following four respects.
\begin{enumerate}
\item	Generalization in learning for goal-directed plan generation; 
\item   Goal-directed plan generation for different types of goals;
\item	Goal understanding from sensory observation for different types of goals;
\item	Rational plan generation.
\end{enumerate}
Following our previous work \cite{Matsumoto_2020}, we first evaluate the generalization capability of the proposed model for reaching untrained goal positions using a limited number of teaching trajectories.
The second test examines how the model can generate goal-directed plans and execute them for different types of goals, in this case reaching specified goal positions and cycling around an obstacle.
The third test demonstrates the capability of the model to infer different types of goals from observed sensation (exteroception).
The fourth test examines the model's capability to generate optimal travel plans to reach specified goals under the constraint of the minimal travel time.

In each test, the simulated agent is in a square workspace with $(x, y)$ coordinates in the range of $[0.0, 1.0]$ for both $x$ and $y$. The agent always starts at position $(0.5, 0.1)$. In the center of the workspace is a fixed obstacle of size $(0.3, 0.05)$. The agent does not directly sense the workspace coordinates. Instead it observes a relative bearing and distance $(\theta_t, \delta_t)$ to a fixed reference point at $(0.0, 0.5)$ as exteroception. The simulated agent controller handles conversion to and from the workspace coordinates to relative bearing-distance as the robot moves. 
At the onset of each test trial, the experimenter sets the preferred goal as a vector $(\hat{g}^\alpha_t, \hat{g}^\beta_t)$. $\hat{g}^\beta_t$ is a three dimensional one-hot vector, with each bucket representing reaching, clockwise cycling, and counter-clockwise cycling goals, respectively. $\hat{g}^\alpha_t$ is set as the goal $x$ coordinate if the reaching goal is set. Otherwise it is left as $0$.
As the network estimates the distal probability $\bar{s}_t$ at each time step, the agent stops at the time step with the maximum estimated distal probability provided that it exceeds a threshold value, assuming that the goal is achieved at that point.

Unless stated otherwise, the experiments have different training data and separately trained networks; however, network parameters are identical between networks. Parameters used for each layer of the RNN are as shown in Table~\ref{tbl:exp1rnn}. Each network was trained for 100,000 epochs, using the Adam optimizer with parameters $\alpha = 0.001$, $\beta_1=0.9$, $\beta_2=0.999$. During planning, the parameters are slightly modified to $\alpha = 0.04$, 500 iterations per sensorimotor time step, and a planning window length of 70. The meta-prior $w$ remains the same in all cases.
% The MDPI table float is called table
\begin{table}[H] % Table 1
\centering
\caption{PV-RNN parameters for~Experiment 1. $\mathbb{R}^d$ and $\mathbb{R}^z$ refer to the number of deterministic ($d$) units and probabilistic ($z$) units respectively. $w_{t=1}$ refers to the meta-prior setting at the first time step (in our previous work this was referred to as $w_I$).} \label{tbl:exp1rnn}
%%% \tablesize{} %% You can specify the fontsize here, e.g., \tablesize{\footnotesize}. If commented out \small will be used.
\begin{tabular}{cccc}
\toprule
                          & \multicolumn{3}{c}{\textbf{Layer}} \\
	                      & \textbf{1}	& \textbf{2} & \textbf{3} \\
\midrule
$\mathbb{R}^d$     	      & 60			& 40         & 20 \\
$\mathbb{R}^z$     	      & 6			& 4          & 2  \\
$\tau$                    & 2           & 4          & 8 \\
$w$                       & 0.0001      & 0.0005     & 0.001 \\
$w_{t=1}$                 & 1.0         & 1.0        & 1.0 \\
\bottomrule
\end{tabular}
\end{table}

%%% Experiment 1A
\subsubsection{Experiment 1A: Generalization in plan generation by learning} \label{sec:exp1a}
\begin{figure}[H] % Fig.7
\centering
\subcaptionbox{\label{fig:exp1a_data_lr}}{\includegraphics[width=0.33\linewidth]{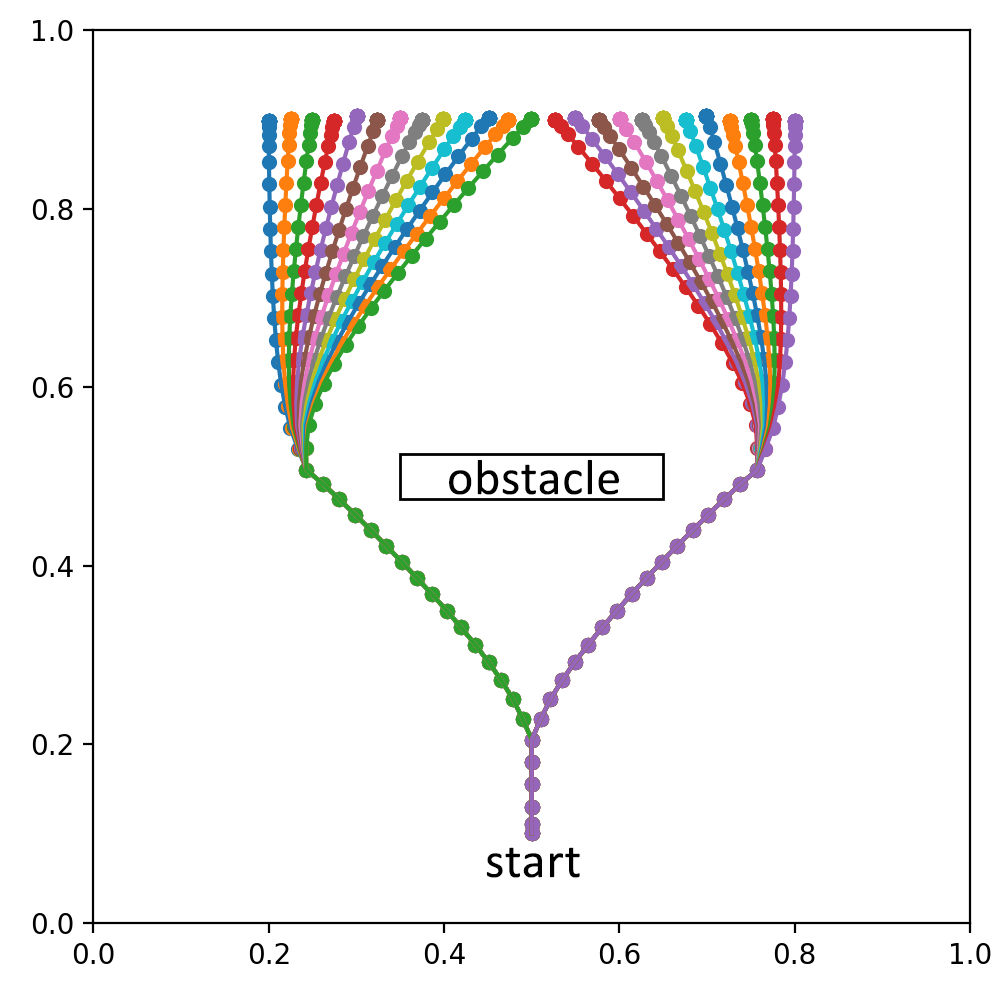}}
\subcaptionbox{\label{fig:exp1a_data_lr2}}{\includegraphics[width=0.33\linewidth]{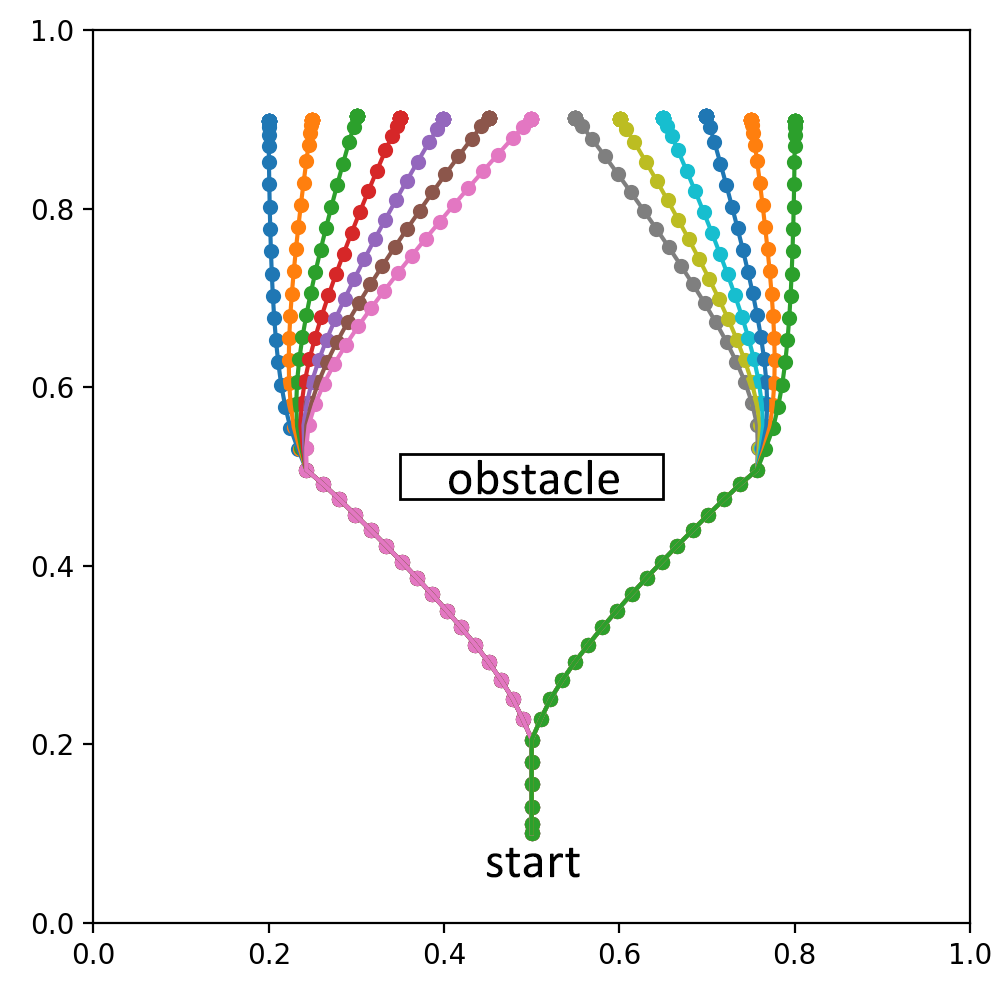}}
\vspace{1em}

\subcaptionbox{\label{fig:exp1a_data_lr4}}{\includegraphics[width=0.33\linewidth]{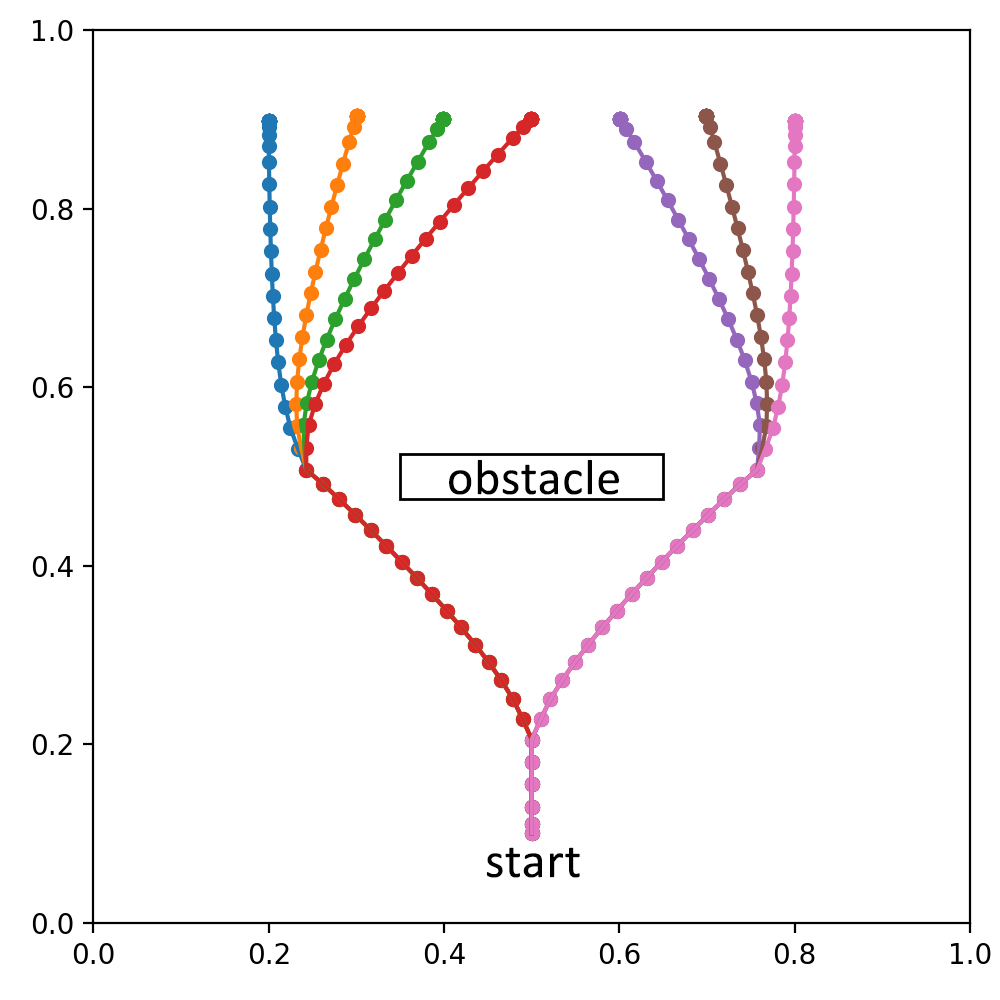}}
\subcaptionbox{\label{fig:exp1a_data_lr8}}{\includegraphics[width=0.33\linewidth]{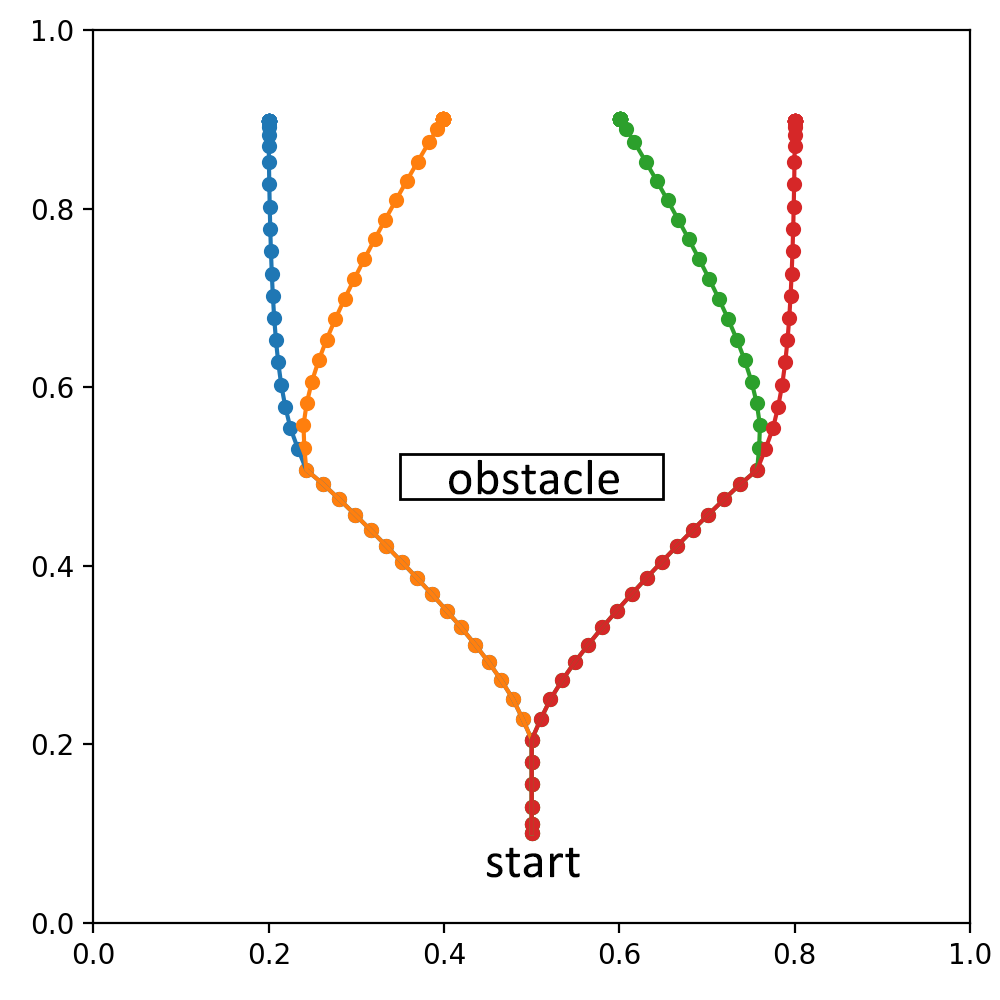}}
\caption{Testing goal generalization by reducing the number of training samples (\textbf{a}) 25 training samples (\textbf{b}) 13 training samples (\textbf{c}) 7 training samples (\textbf{d}) 4 training samples}
\label{fig:exp1a_data}
\end{figure}

In order to evaluate how well the network can generalize goal positions in the goal area with a limited number of teaching trajectories, we prepared four teaching datasets with decreasing numbers of goal locations to be reached as shown in Figure~\ref{fig:exp1a_data}. 
The goal locations are on a line in the range $x=[0.2,0.8]$, $y=0.9$. The agent accelerates up to speed to a branching point, and then either turns left or right to the side of the obstacle, before moving toward the goal position. As the agent approaches the goal position, it decelerates to a stop.

The maximum trajectory length is 70, although the distal step occurs at around 40 time steps. To account for randomness in training, five networks with different initial random seeds are trained for each of the four datasets, for 20 trained networks in total. Untrained test goals were drawn from a uniformly random distribution in the range $[0.2, 0.8]$. Each network was tested with ten untrained goals, with the results averaged over all test goals and networks for each dataset.

\begin{figure}[H] % Fig. 8
\centering
\includegraphics[width=0.6\linewidth]{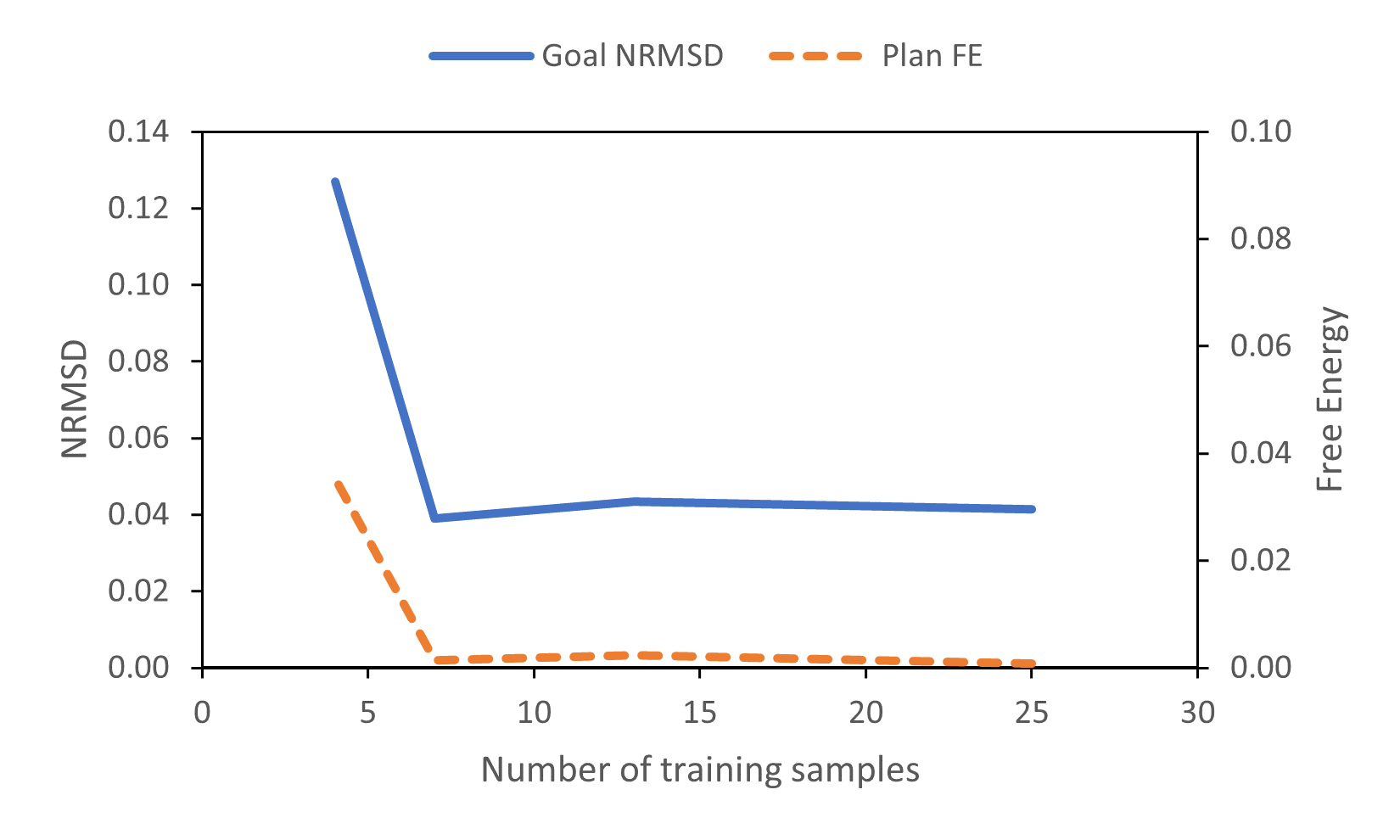}
\caption{Comparison of goal deviation and residual free energy for different numbers of training goals.}
\label{fig:exp1a_result}
\end{figure}

To evaluate goal generalization, we considered the difference between the final agent position reached at the end of plan execution and the preferred goal at the end for each test trial, as well as the plan free energy that remained. The difference between agent position and goal is expressed as the root-mean-square deviation, normalized to the range of $\bar{g}^\alpha$ (NRMSD). The result is summarized in Figure~\ref{fig:exp1a_result}.

We observed that the network achieved stable goal-position generalization when at least 7 training trajectories are used. 
Also, it can be seen that the plan free energy was minimized in a similar manner.

%%% Experiment 1B
\subsubsection{Experiment 1B: Goal-directed plan generation for different types of goals} \label{sec:exp1b}
\begin{figure}[H] % Fig. 9
\centering
\subcaptionbox{\label{fig:exp1b_data_alr}}{\includegraphics[width=0.33\linewidth]{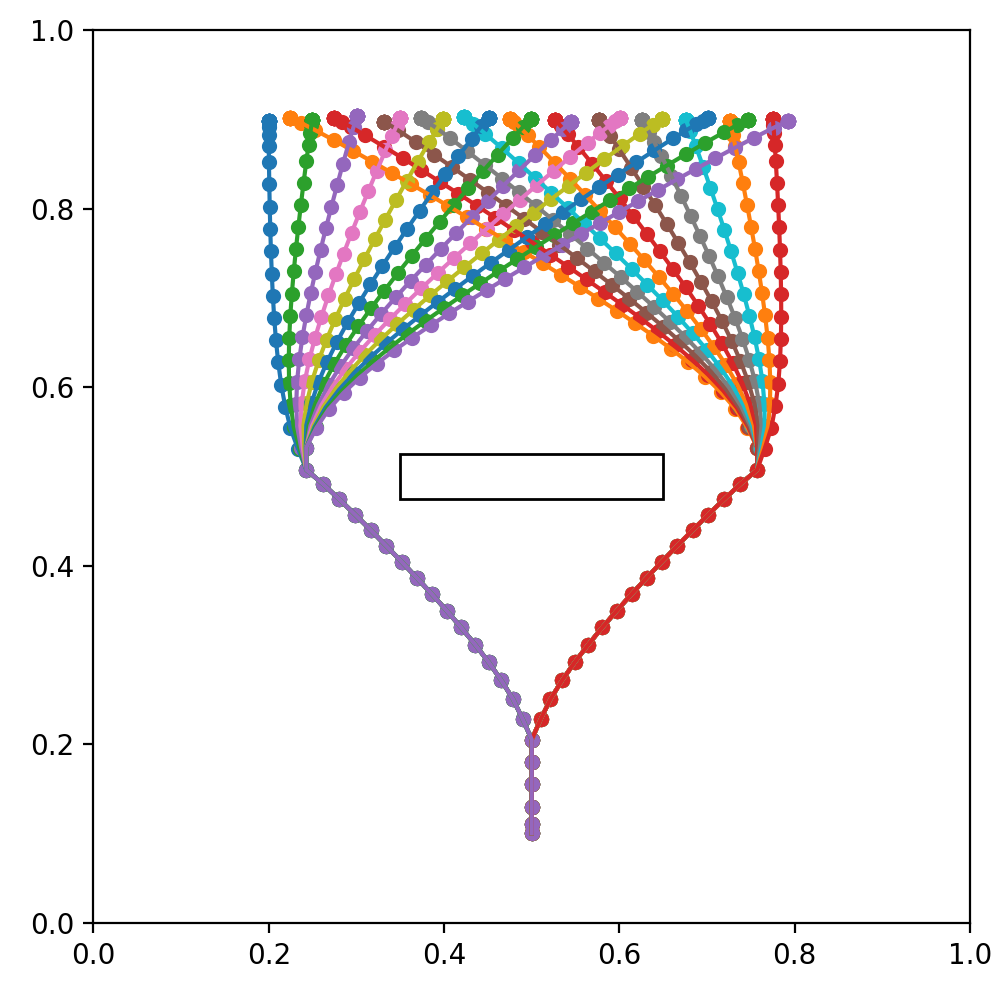}}
\subcaptionbox{\label{fig:exp1b_data_ao}}{\includegraphics[width=0.33\linewidth]{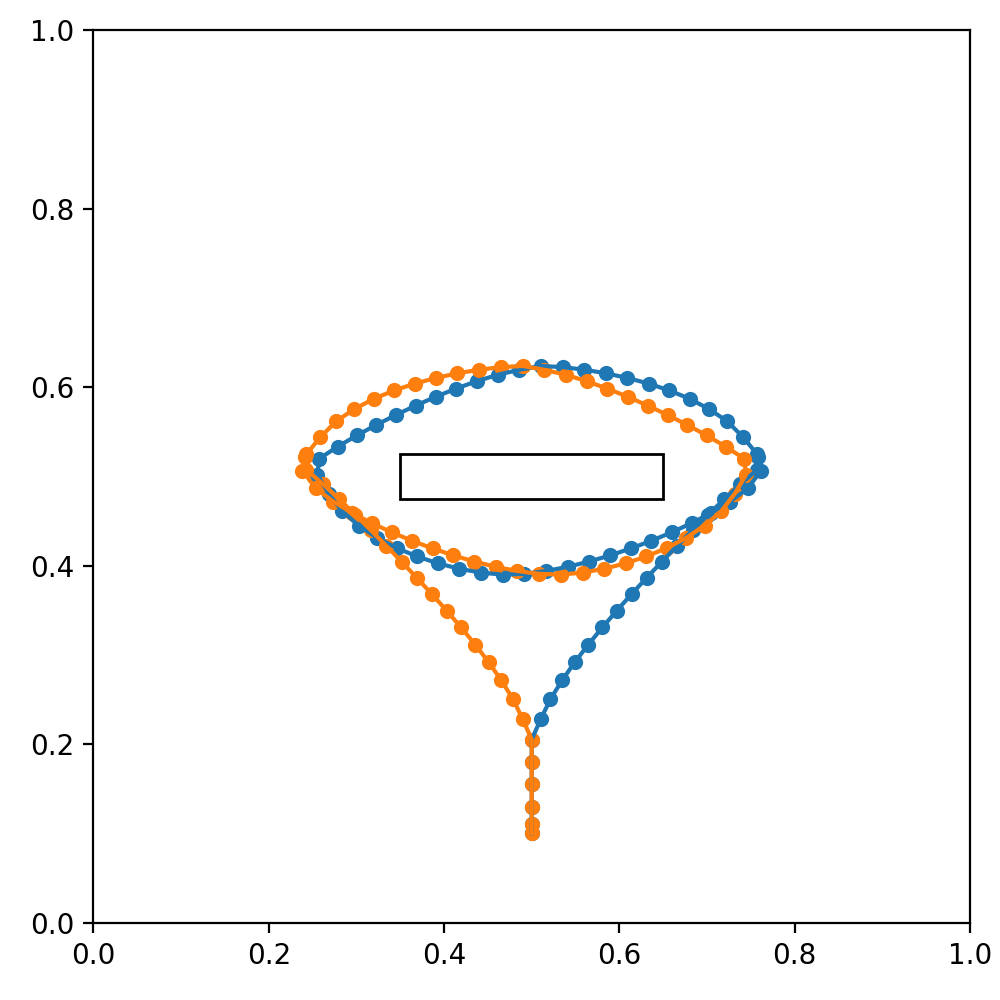}}
\caption{Teaching trajectories used for Experiment 1B, with both goal reaching and cycling goals. (\textbf{a}) 25 reaching trajectories, with both short and long paths (\textbf{b}) 2 cycling trajectories, in the clockwise and counter-clockwise directions}
\label{fig:exp1b_data}
\end{figure}

For this test, we used a more complex set of teaching trajectories, containing three distinct goals with significantly different patterns. The first goal, shown in Figure~\ref{fig:exp1b_data_alr}, is similar to the previous goal-reaching trajectories shown in Figure~\ref{fig:exp1a_data_lr}; however, this scenario is ill-posed. That is, the trajectories alternate between short and long paths for the same goal position. This set of teaching trajectories will also be reused in Experiment 1C and Experiment 1D. We note that while the training data themselves are not ill-posed, due to generalization, the learning outcome is ill-posed. We will revisit this issue in Section~\ref{sec:exp1d}.
The second and the third goals, consisting of the two training trajectories shown in Figure~\ref{fig:exp1b_data_ao}, involve the agent cycling around the central obstacle in a clockwise direction and a counter-clockwise direction, respectively. 
For the reaching goal, once the agent reaches the goal position, the distal step is set and the remaining steps are padded with the final value, i.e., the agent remains stationary.
Unlike the reaching goal, the cycling goals have no distal step. The training sequence demonstrates a single cycle; however, it is desirable for the agent to continue the action indefinitely. There are a total of 27 teaching trajectories in this training dataset, each of which has a length of 70 time steps.

\begin{figure}[H] % Fig.10
\centering
\subcaptionbox{\label{fig:exp1b_results_pgen_gr}}{\includegraphics[width=0.8\linewidth]{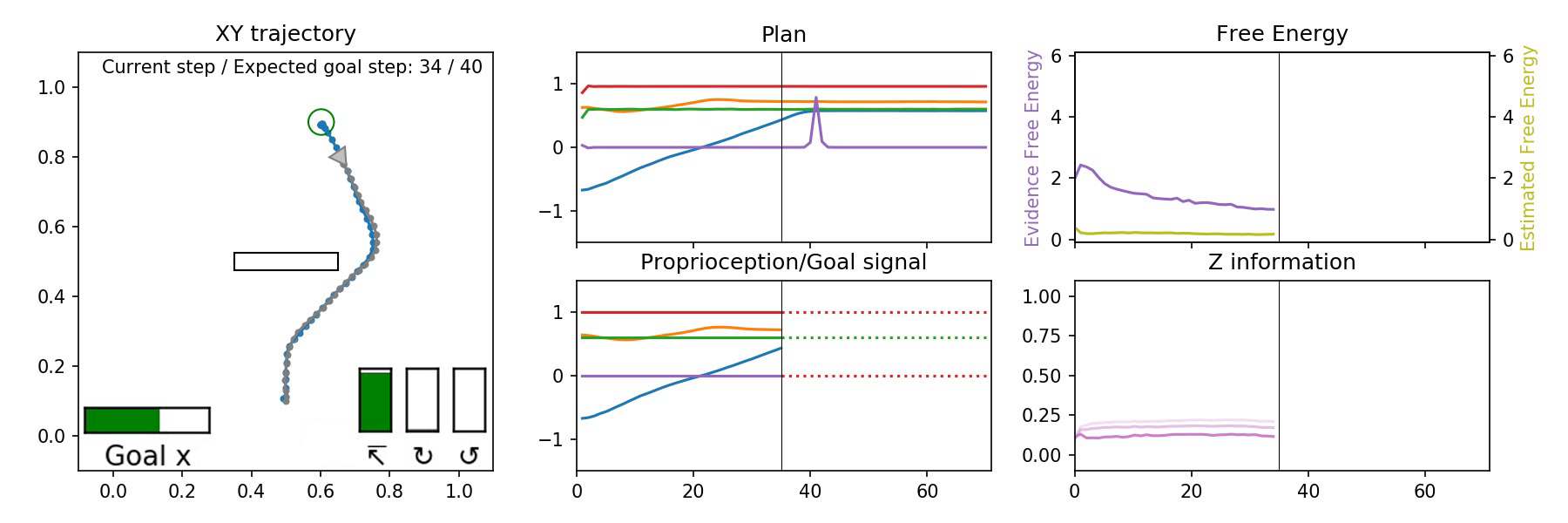}}

\vspace{1em}

\subcaptionbox{\label{fig:exp1b_results_pgen_cw}}{\includegraphics[width=0.8\linewidth]{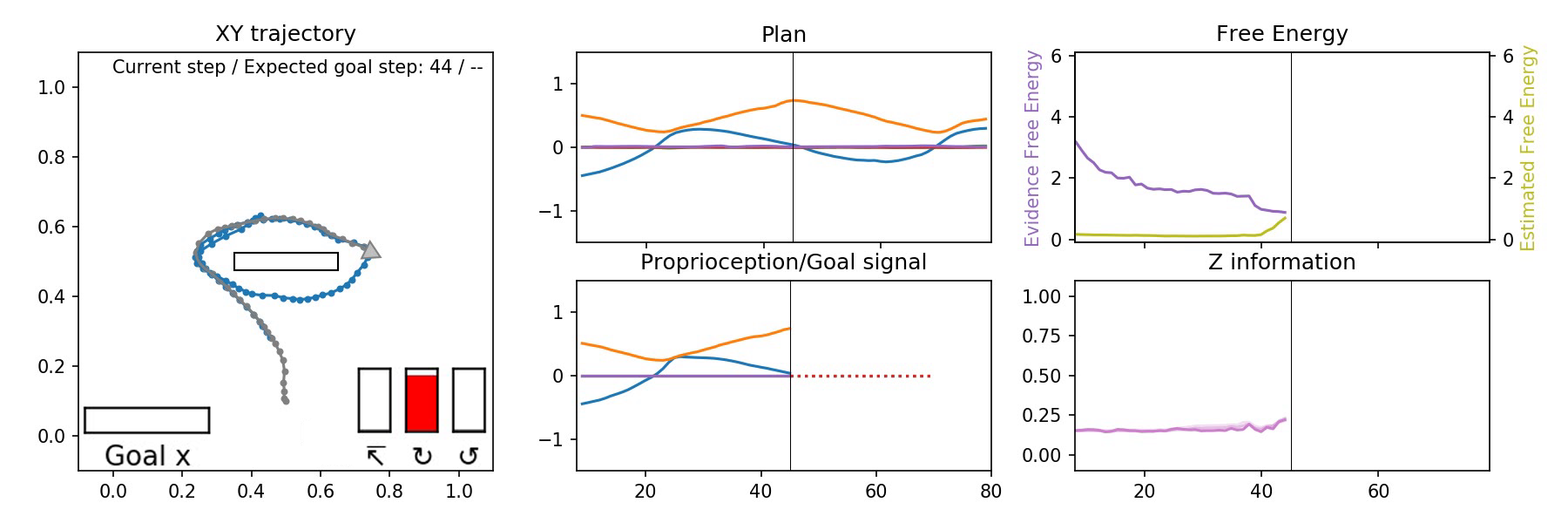}}

\vspace{1em}

\subcaptionbox{\label{fig:exp1b_results_pgen_ccw}}{\includegraphics[width=0.8\linewidth]{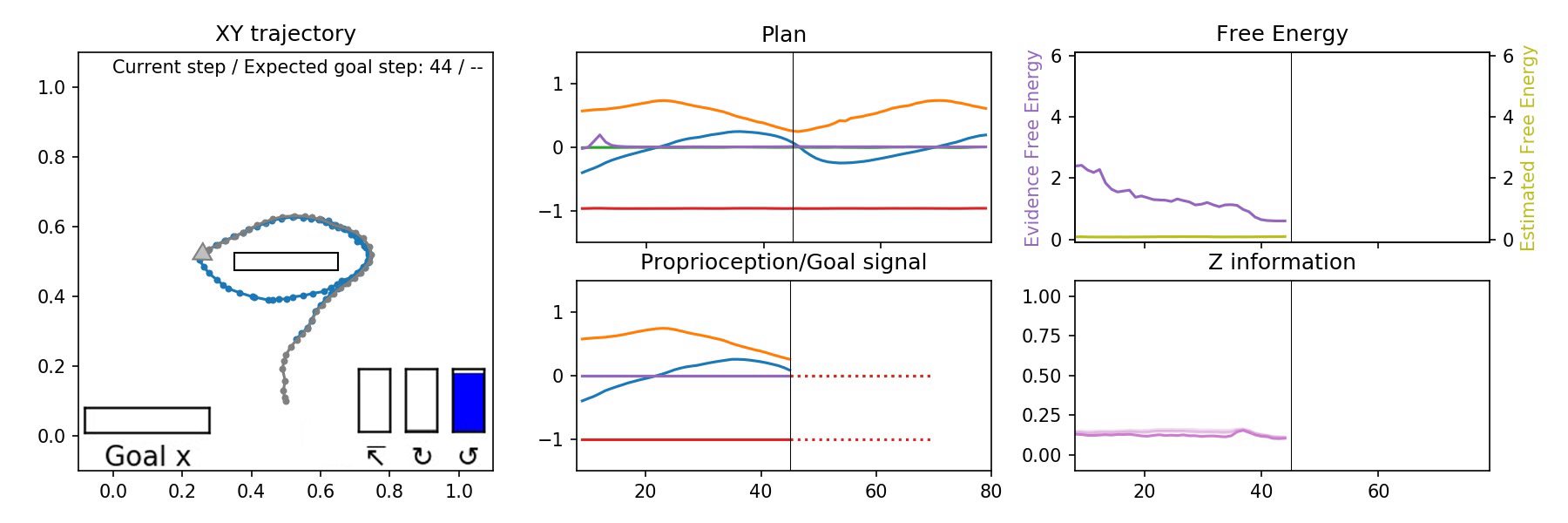}}
\caption{Examples of trajectories generated for three different goal categories, (\textbf{a}) reaching, (\textbf{b}) clockwise cycle, and (\textbf{c}) counter-clockwise cycle from the results of Experiment 1B. The left column shows a view of the 2D workspace, with the generated plan trajectory, shown as a solid line with dots at each time step and the goal position if available. The agent is represented by a triangle, with the agent's position history trailing behind it and overlaid on the plan trajectory.The subset along the bottom (enlarged for visibility) is the expected goal, the left-most horizontal bar is $\bar{g}^\alpha$, while the three vertical bars is $\bar{g}^\beta$; the left vertical bar represents reaching, the middle bar represents clockwise and the right bar represents counter-clockwise goals. The height of the vertical bar represents the probability (confidence) of this goal. The distal step, if available, is also shown both as a circle in 2D and in text in the top right. 
The middle column shows the plan in terms of exteroceptive trajectory in the top as well as the observed exteroception trajectory and preferred goal (encoded for display as scalar values 1, 0 and -1 for reaching, clockwise and counter-clockwise respectively) in dotted lines in the bottom, with the vertical black bar representing the current sensorimotor time step. The right column shows the evidence free energy (dark purple) and the expected free energy (light green) in the top and Z information, a measure of how much of a contribution is made by the probabilistic (z) units in each layer (darker lines are higher layers), equal to $D_{KL}[q || N]$ where $N$ is the unit normal distribution.}
\label{fig:exp1b_results_pgen}
\end{figure}

We evaluated how precisely the agent can generate movement trajectories for achieving goals specified. 
In Table~\ref{tbl:exp1b_results_pgen} we summarize the results for the reaching and cyclic goals. For the reaching goal, as used previously, average NRMSD is given for 10 random untrained goal positions. For the cyclic goal, NRMSD between the agent's movement and the training sequence for the entire 70 time step sequence is taken, and averaged for five clockwise and counter-clockwise orbits each. In the latter case, normalization is done over the entire dataset range rather than the goal range.
Table~\ref{tbl:exp1b_results_pgen} confirms that both types of goals can be achieved within minimal range of goal error.
\begin{table}[H] 
\centering
\caption{Deviation from the ground truth, given as normalized RMS.} \label{tbl:exp1b_results_pgen}
%%% \tablesize{} %% You can specify the fontsize here, e.g., \tablesize{\footnotesize}. If commented out \small will be used.
\begin{tabular}{ccc}
\toprule
	              & \textbf{Reaching} & \textbf{Cycling} \\
\midrule
NRMSD     	      & 0.033241		  & 0.028158 \\
\bottomrule
\end{tabular}
\end{table}
Figure~\ref{fig:exp1b_results_pgen} shows three examples of trajectories generated for three goal categories, with trajectories of motion in the left panel, the representative network values in the middle, and the expected free energy for the past window and the evidence free energy in the future window in the right panel. 
We observed that plan generation stabilized quickly after the onset of travel by minimizing the expected free energy, and the network was able to accurately estimate the distal step in the case of reaching the goal. 
Although the future plan trajectory was constantly changing due to the scheme of online plan generation and the error regression in the past, the motor plan generated was stably executed at each sensorimotor time step.
It can be also observed that the evidence free energy in the past window continued to converge in all three plots, meaning that all latent variables were gradually situated to the behavioral context.
Therefore, the deviation of the executed trajectory from the planned trajectory was limited.
The full observed temporal processes can be seen in the recorded videos of the simulations at this \href{https://youtu.be/9aJYkmzDJts}{link}.

%%% Experiment 1c
\subsubsection{Experiment 1C: Goal inference by sensory observation} \label{sec:exp1c}
\begin{figure}[H] % Fig.11
\centering
\subcaptionbox{\label{fig:exp1c_data_target_lg}}{\includegraphics[width=0.33\linewidth]{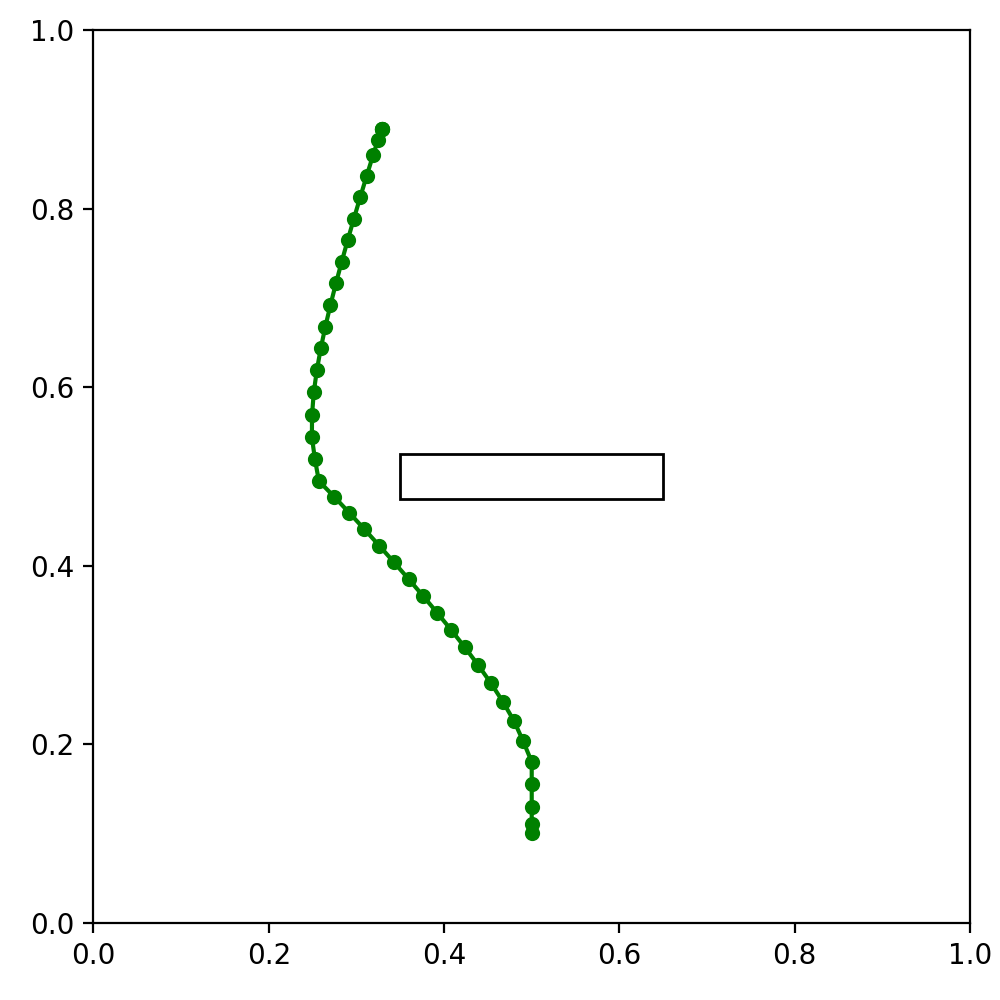}}
\subcaptionbox{\label{fig:exp1c_data_target_rg}}{\includegraphics[width=0.33\linewidth]{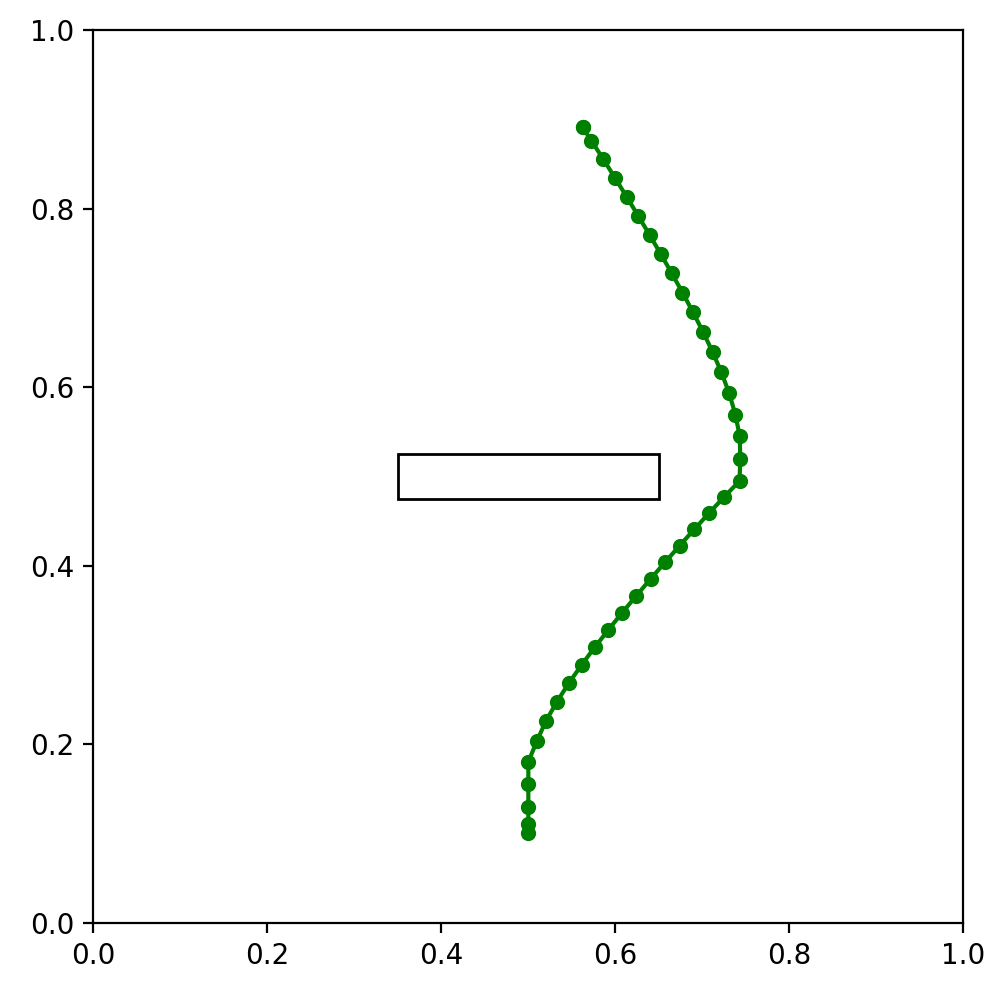}}

\vspace{1em}

\subcaptionbox{\label{fig:exp1c_data_target_cw}}{\includegraphics[width=0.33\linewidth]{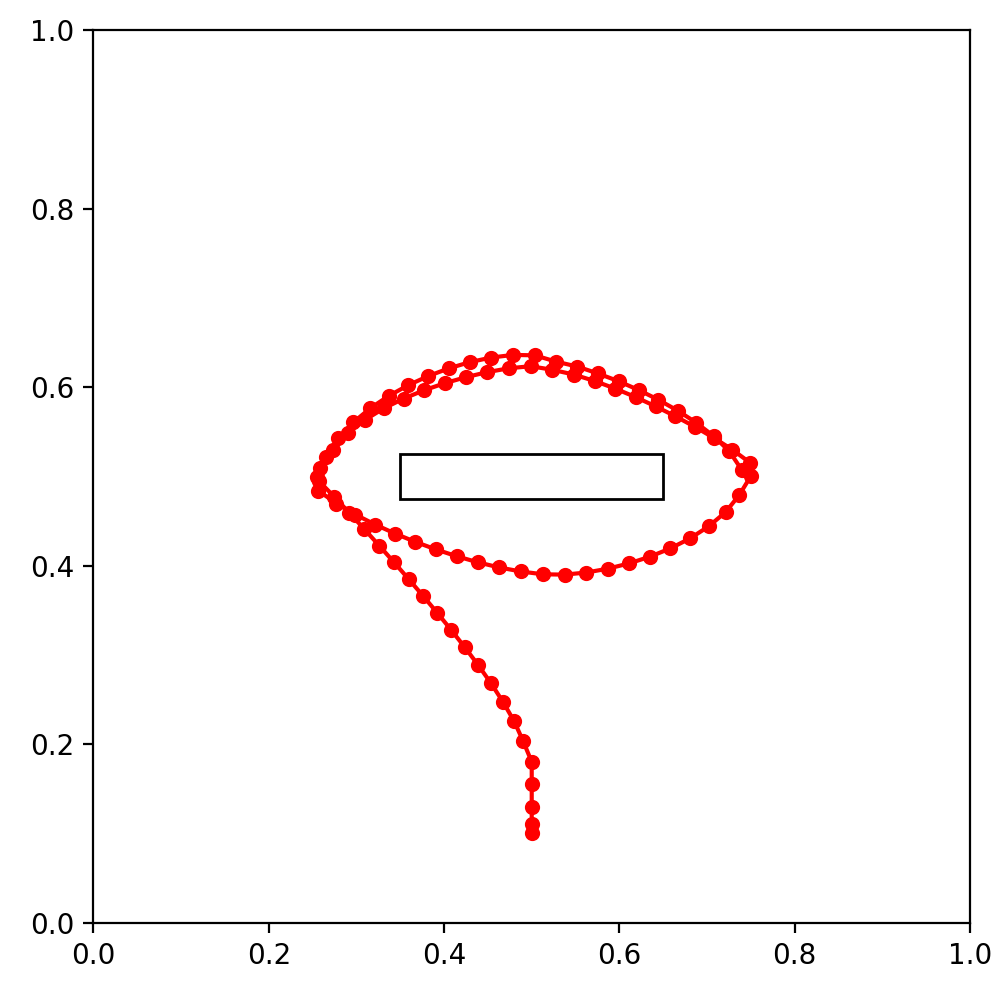}}
\subcaptionbox{\label{fig:exp1c_data_target_ccw}}{\includegraphics[width=0.33\linewidth]{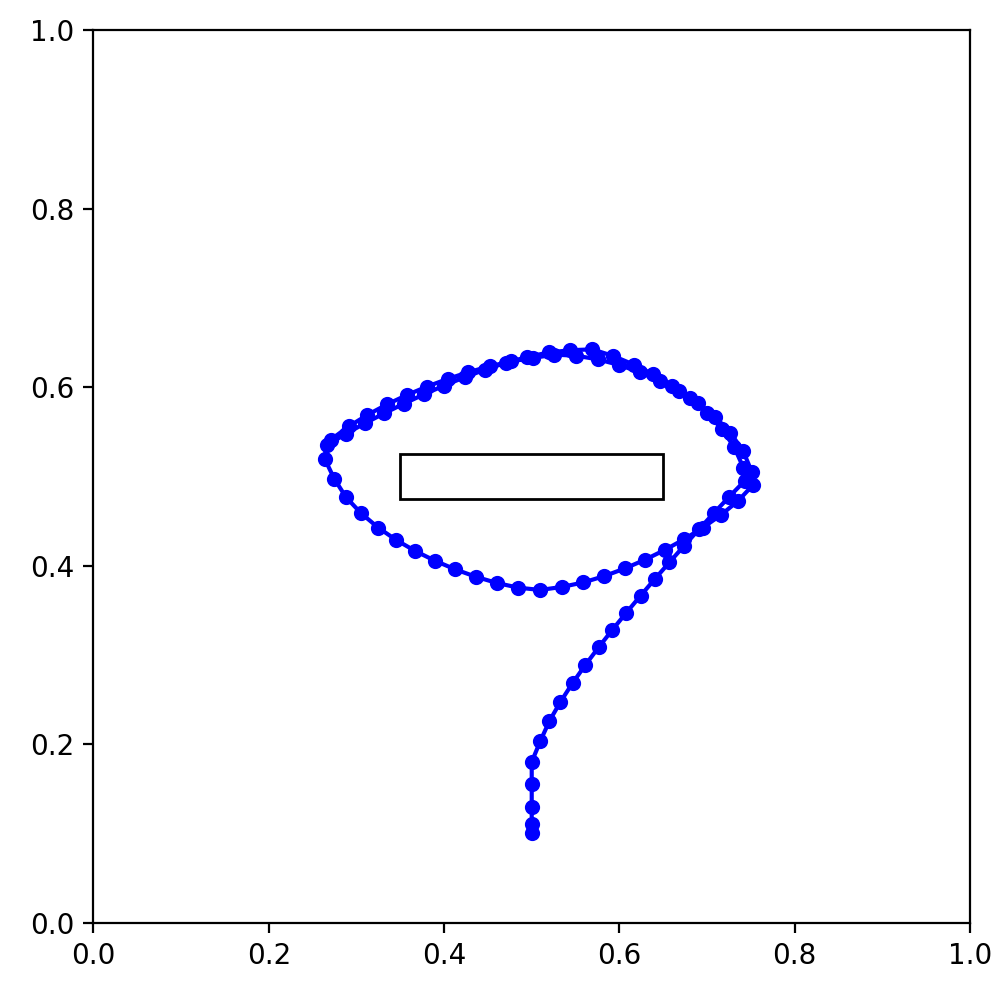}}
\caption{Test trajectories used for goal inference, color coded by the type of goal. These trajectories represent agent actions for (\textbf{a}) reaching goal located on the left, (\textbf{b}) reaching goal located on the right, (\textbf{c}) clockwise cycling, and (\textbf{d}) counter-clockwise cycling. 
The reaching trajectories and the cycling trajectories are 40 and 90 steps long, respectively.}
\label{fig:exp1c_data_targets}
\end{figure}

\begin{figure}[H] % Fig.12
\subcaptionbox{\label{fig:exp1c_results_gu_grl1}}{\includegraphics[width=0.33\linewidth]{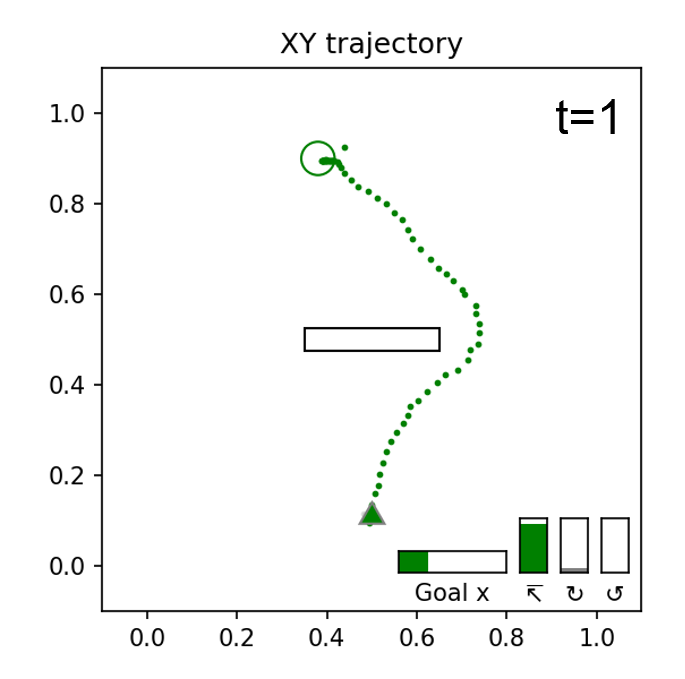}}
\subcaptionbox{\label{fig:exp1c_results_gu_grl12}}{\includegraphics[width=0.33\linewidth]{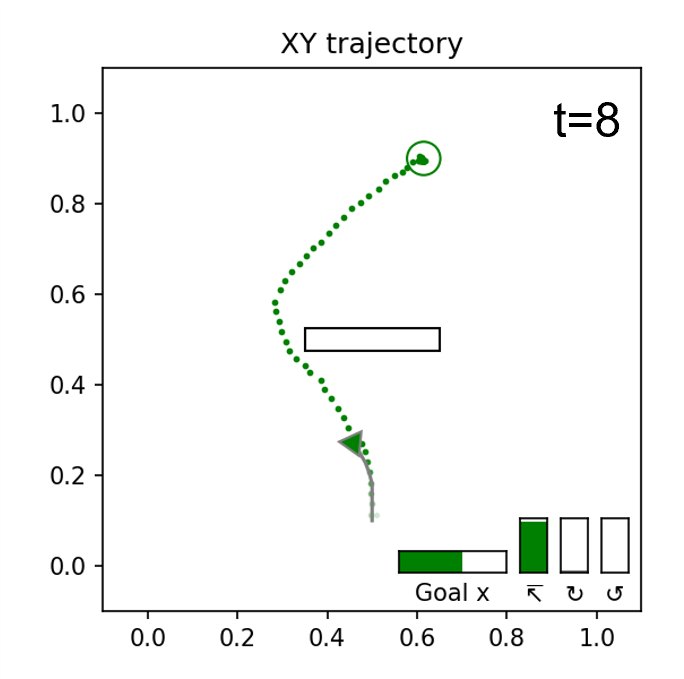}}
\subcaptionbox{\label{fig:exp1c_results_gu_grl29}}{\includegraphics[width=0.33\linewidth]{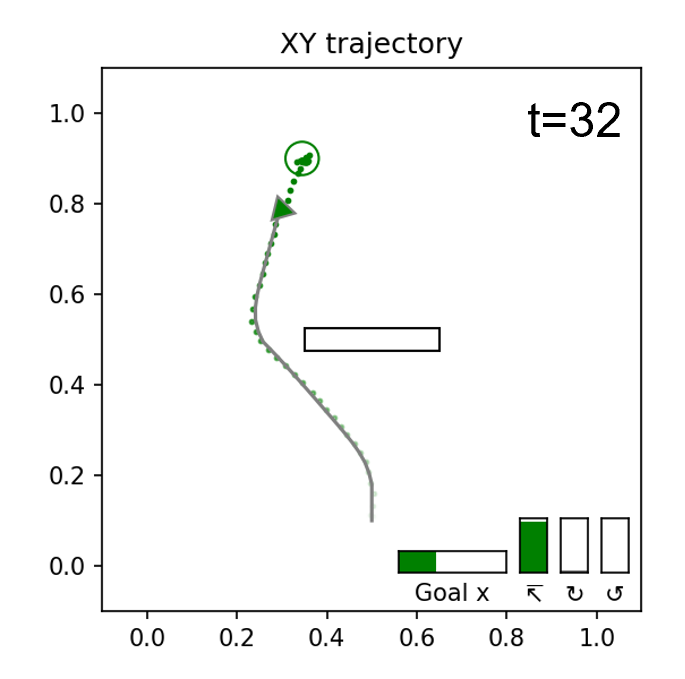}}

\vspace{1em}

\subcaptionbox{\label{fig:exp1c_results_gu_grr1}}{\includegraphics[width=0.33\linewidth]{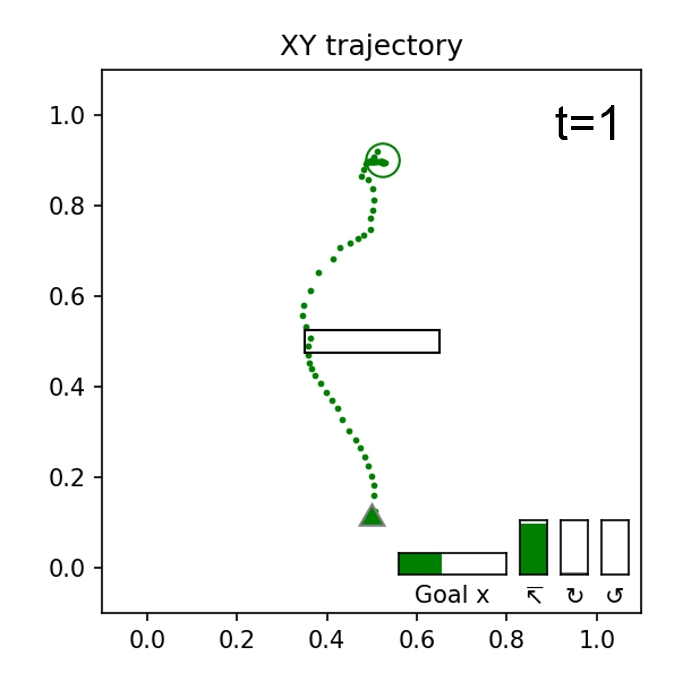}}
\subcaptionbox{\label{fig:exp1c_results_gu_grr7}}{\includegraphics[width=0.33\linewidth]{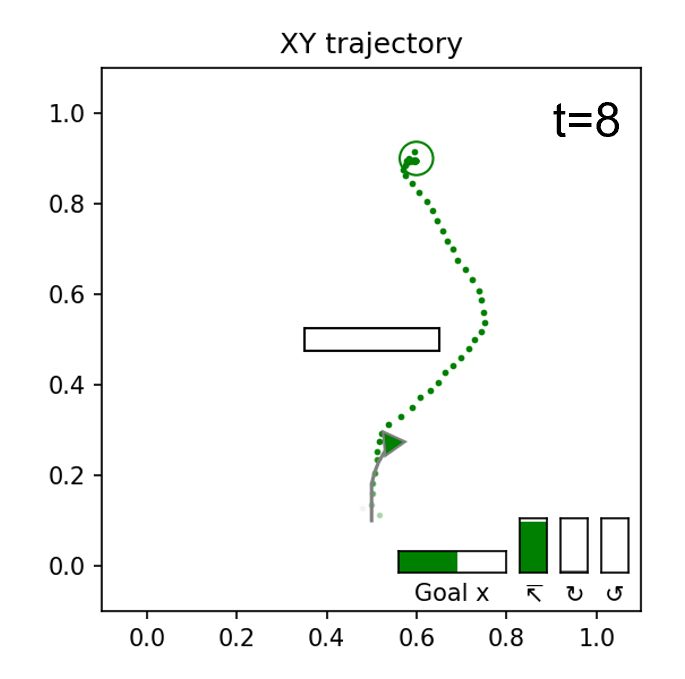}}
\subcaptionbox{\label{fig:exp1c_results_gu_grr29}}{\includegraphics[width=0.33\linewidth]{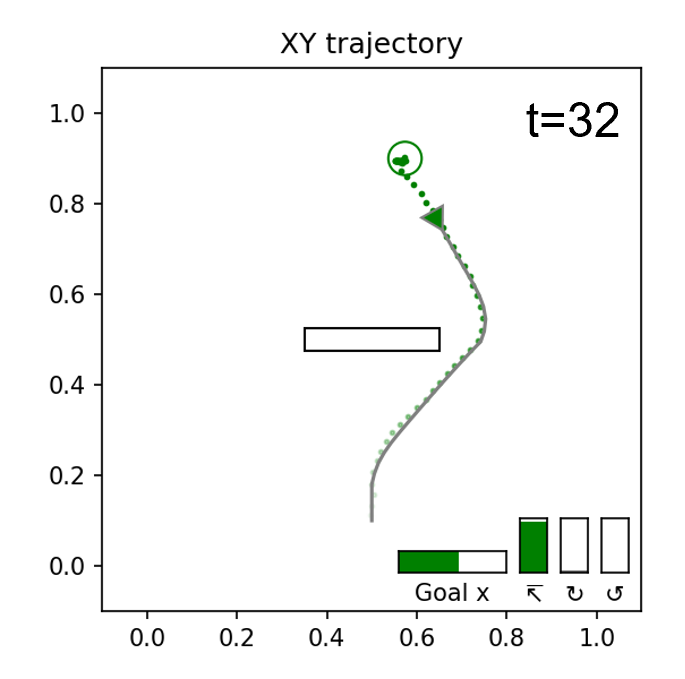}}
\caption{Anticipated goal and trajectories as the network observes the reaching trajectories at different time steps during the observation of travels. The colored dots represent the anticipated trajectory, while the solid gray line is the observed trajectory. The time step is shown in the top right. (\textbf{a} -- \textbf{c}) Agent observing the trajectory reaching to the left goal, (\textbf{d} -- \textbf{f}) agent observing the trajectory reaching to the right goal.}
\label{fig:exp1c_results_gu1}
\end{figure}

Next, we evaluated how precisely the network trained in Experiment 1B can infer goals as well as future movement trajectories by observing movement trajectories in terms of the exteroception sequence.
In this experiment, four movement trajectories achieving different goals are prepared, which reach goals located left and right of the goal line, clockwise cycling, and counter-clockwise cycling.
The four test trajectories, shown in Figure~\ref{fig:exp1c_data_targets}, were generated in the same way as the training data, but with untrained goal positions for the reaching goals and with a slight variation and extended length for the cycling goals.
Figure~\ref{fig:exp1c_results_gu1} shows the anticipated trajectories and goals at different points in time as the network observed the goal-reaching trajectories in Figures~\ref{fig:exp1c_data_target_lg} and \ref{fig:exp1c_data_target_rg}. 
Initially, at $t=1$ (Figures \ref{fig:exp1c_results_gu_grl1} \& \ref{fig:exp1c_results_gu_grr1}), before the agent observes any movement, the network makes a guess based on the learned prior distribution. 
Since the goal-reaching trajectories are most frequent in the teaching trajectory distribution, reaching a goal is inferred as a default goal when observing the movement trajectory at the starting point. 
Several steps later, at around $t=8$ (Figures \ref{fig:exp1c_results_gu_grl12} \& \ref{fig:exp1c_results_gu_grr7}), the movement trajectory branches either left or right around the obstacle.
We observed that in the case of the goal located on the left, the network initially anticipates a longer path going around the right side of the obstacle before observing that the movement trajectory goes around the other side of the obstacle.
The observed phenomenon is due to the fact that the teaching trajectories contain two possible paths reaching the same goal position.
Therefore, the network can generate two possible movement trajectory plans in the current ill-posed setting.
This issue will be revisited in Section~\ref{sec:exp1d}. 
The anticipated goal coordinate $\bar{g}^\alpha$ is refined as the movement trajectory approaches the goal area. By $t=32$ (Figures \ref{fig:exp1c_results_gu_grl29} \& \ref{fig:exp1c_results_gu_grr29}), the goal is fully anticipated.

\begin{figure}[H] % Fig.13
\subcaptionbox{\label{fig:exp1c_results_gu_grccw22}}{\includegraphics[width=0.33\linewidth]{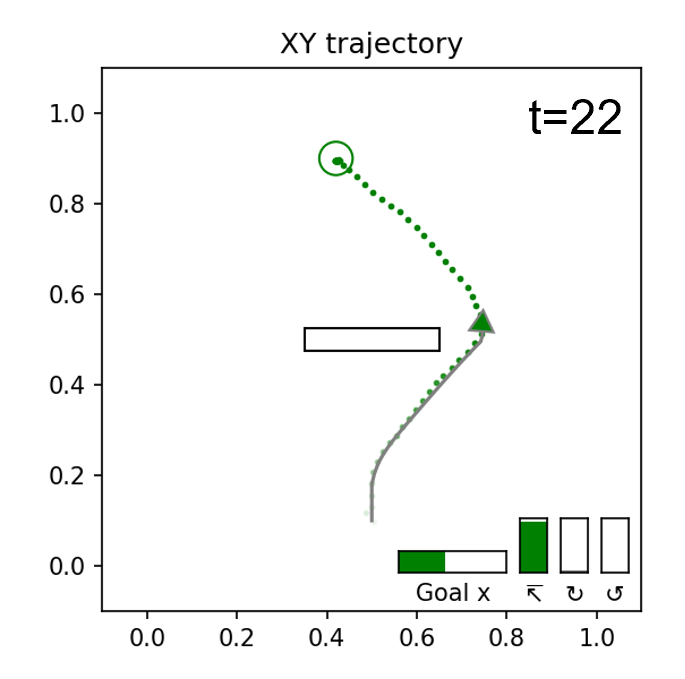}}
\subcaptionbox{\label{fig:exp1c_results_gu_grccw34}}{\includegraphics[width=0.33\linewidth]{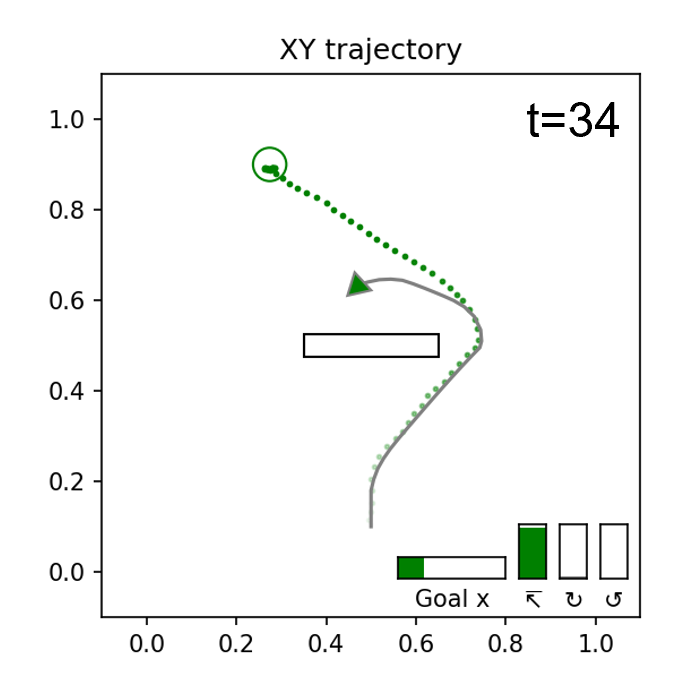}}
\subcaptionbox{\label{fig:exp1c_results_gu_grccw37}}{\includegraphics[width=0.33\linewidth]{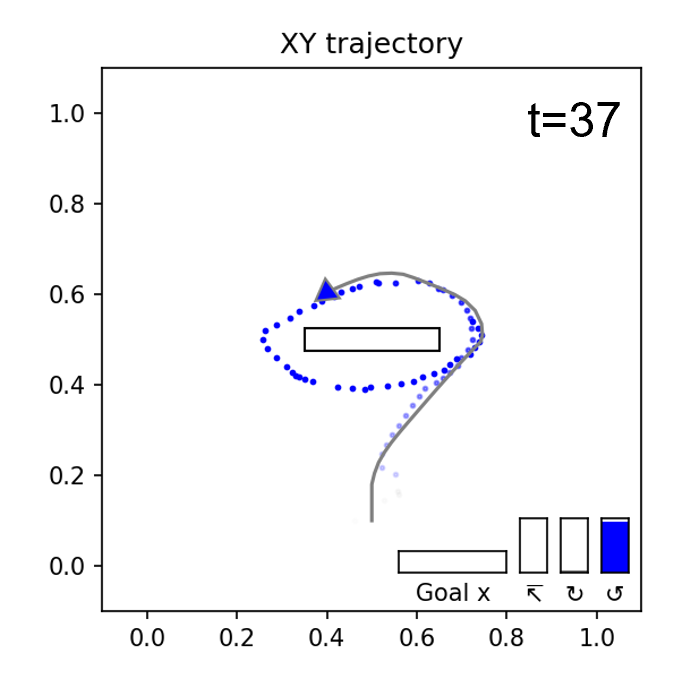}}

\vspace{1em}

\subcaptionbox{\label{fig:exp1c_results_gu_grcw22}}{\includegraphics[width=0.33\linewidth]{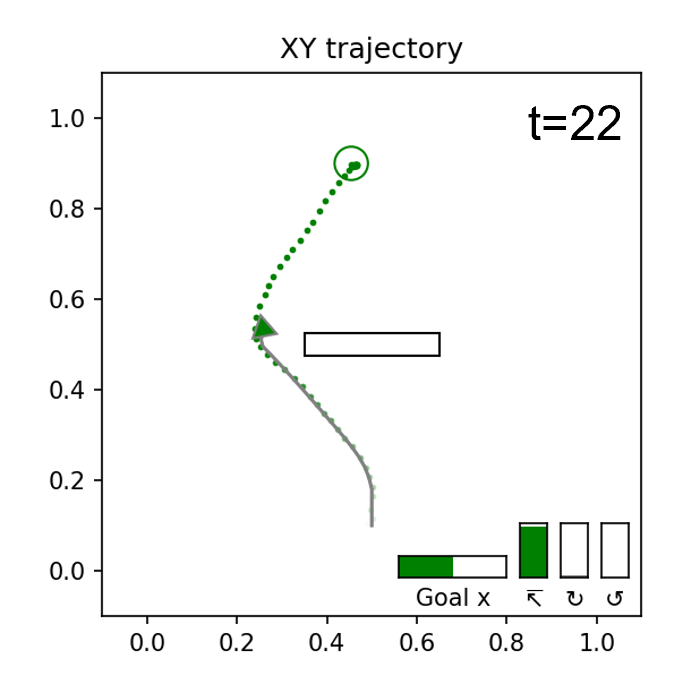}}
\subcaptionbox{\label{fig:exp1c_results_gu_grcw49}}{\includegraphics[width=0.33\linewidth]{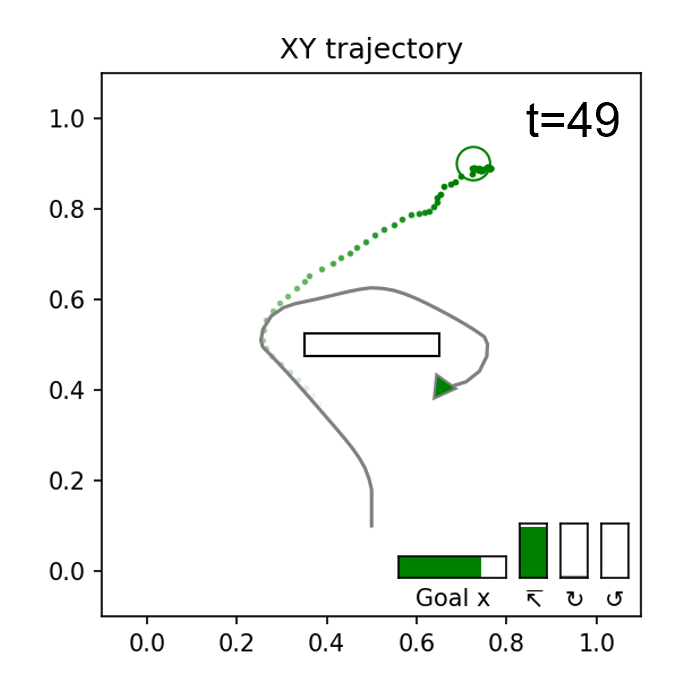}}
\subcaptionbox{\label{fig:exp1c_results_gu_grcw79}}{\includegraphics[width=0.33\linewidth]{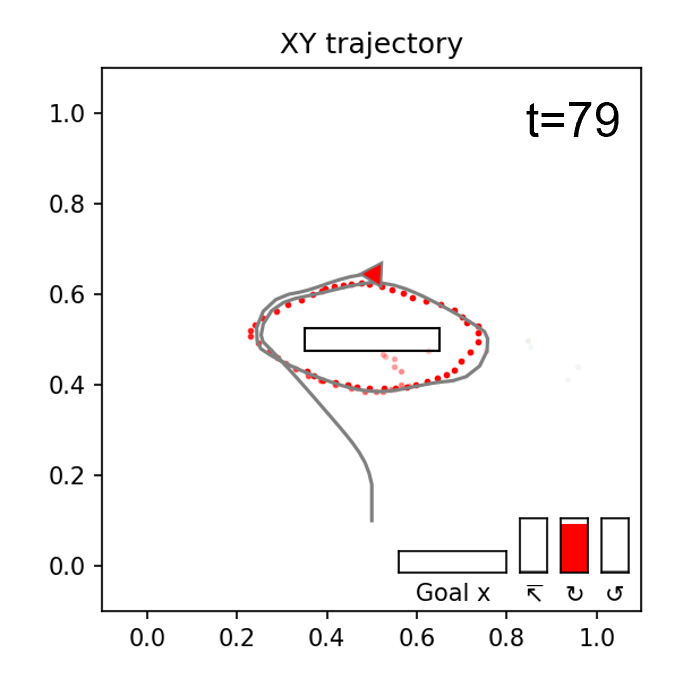}}
\caption{Anticipated goal and trajectories as the network observes the cyclic trajectories. The colored dots represent the anticipated trajectory, while the solid gray line is the observed trajectory. The time step is shown in the top right. (\textbf{a} -- \textbf{c}) Agent following the counter-clockwise cyclic goal trajectory, (\textbf{d} -- \textbf{f}) agent following the clockwise cyclic goal trajectory.}
\label{fig:exp1c_results_gu2}
\end{figure}

\begin{figure}[H] % Fig.14
\centering
\includegraphics[width=0.8\linewidth]{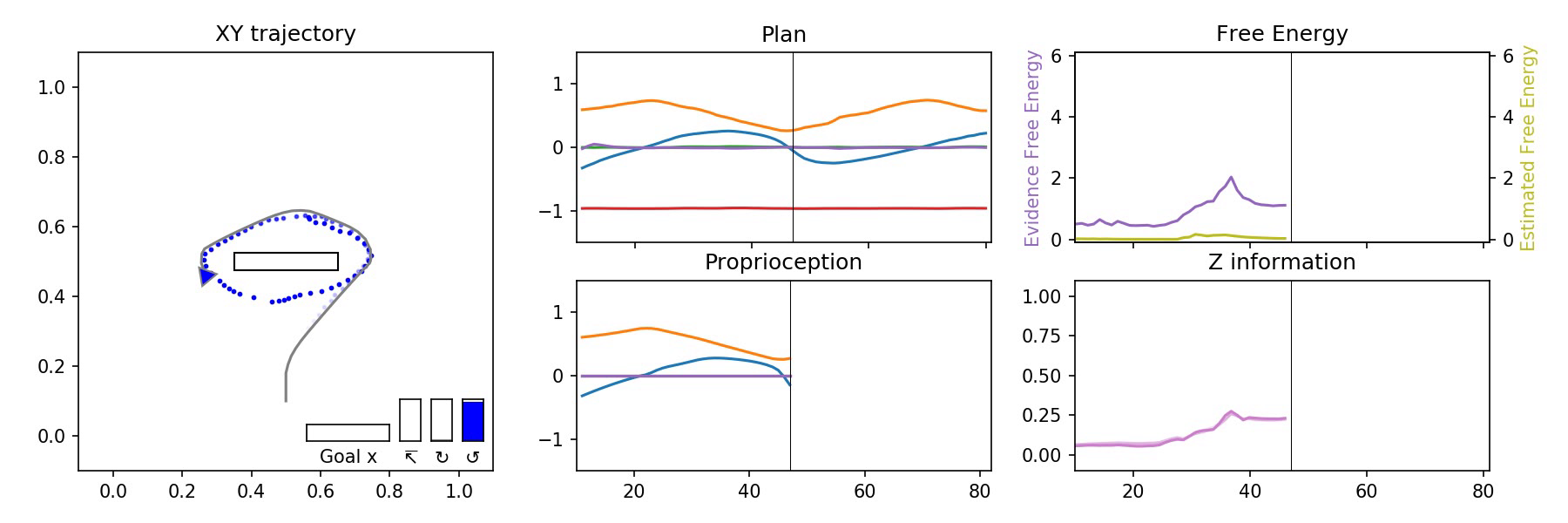}
\caption{Network inferring the counter-clockwise cycling goal while observing the movement trajectory in the left panel. The center panel shows the inferred plan in the top and the observed exteroception in the bottom. Note the peak in the free energy before the goal is correctly inferred in the right panel.}
\label{fig:exp1c_results_gu_grccw}
\end{figure}

Figure~\ref{fig:exp1c_results_gu2} shows the trajectories and goals inferred for the observed cyclic trajectories shown in Figures~\ref{fig:exp1c_data_target_cw} and \ref{fig:exp1c_data_target_ccw}. 
Until $t=22$ the observed trajectories are mostly indistinguishable from the reaching trajectories, of which the occurrence probability was learned as high in the prior distribution, the network infers goal reaching as the default goal from these observations.
While the observed movement trajectory begins to enter a cyclic trajectory after $t=22$, it still takes some time for the network to correctly infer the ongoing goal as cyclic. 
During this time, free energy increases, an example of which is shown in Figure~\ref{fig:exp1c_results_gu_grccw}. 
This is analogous to a `surprising' observation. 
After some time, the goals are inferred correctly as the cycling goals as shown in Figures~\ref{fig:exp1c_results_gu_grccw37} and \ref{fig:exp1c_results_gu_grcw79} for the counter-clockwise and clockwise cases, respectively.
The free energy is reduced accordingly at this moment, as also shown in Figure~\ref{fig:exp1c_results_gu_grccw}, which shows a generated plan, the plan free energy, and z information in the counter-clockwise case. 
We note that the activity of z units (z information) also rises and stays elevated to contribute to producing the correct inference of possible goals.

We assume that the agent can recognize consecutive switching of goals from observation; thus, we also tested a scenario wherein the agent first observes a clockwise cycling trajectory, then a counter-clockwise trajectory, and a goal-reaching trajectory, all in one continuous sequence.
This is a challenging test, since the network was not trained to cope with such dynamic goal switching. 
A video of this experimental result is provided at this \href{https://youtu.be/x9f1UBkTKqQ}{link}. In the animation it can be seen that the network inferred the changing goals successfully. 
However, the anticipated future trajectories were quite unstable toward the end of the trial.
As we observed that free energy becomes quite large, it is presumed that the goal inference from the observation may take a relatively long time for convergence of the free energy for unlearned situations.
Therefore, it may be difficult for the network to catch up to the goal switching if it occurs too frequently. 

% Experiment 1D
\subsubsection{Experiment 1D: Goal-directed planning enforcing the well-posed condition} \label{sec:exp1d}
\begin{figure}[H] % Fig.15
\centering
\subcaptionbox{\label{fig:exp1d_results_spath}}{\includegraphics[width=0.33\linewidth]{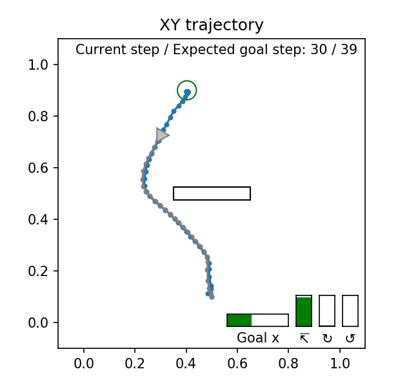}}
\subcaptionbox{\label{fig:exp1d_results_lpath}}{\includegraphics[width=0.33\linewidth]{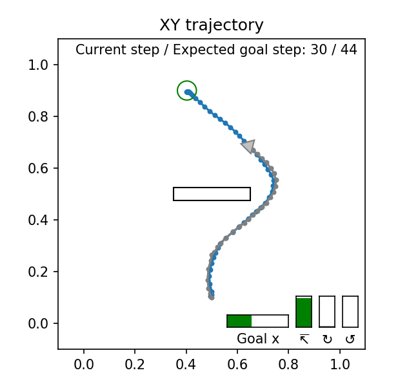}}
\caption{Examples of ill-posed goal-directed planning. (\textbf{a}) Generation of a short path reaching the goal (39 time steps) and (\textbf{b}) an alternative long path to the same goal (44 time steps).}
\label{fig:exp1d_results_slpath}
\end{figure}

In Experiment 1B, the network was trained using teaching trajectories that included alternative trajectories reaching similar goal positions as shown in Figure~\ref{fig:exp1b_data_alr}.
This made the goal-directed planning ill-posed since the network cannot determine an optimal plan between two possible choices under the current definition of the expected free energy.
Figure~\ref{fig:exp1d_results_slpath} shows an illustrative example as the result of the ill-posed goal-directed planning where we see that both a short and long path can be generated for the same goal.
Conventionally, it has been shown that the problem of ill-posed, goal-directed planning can be transformed into a well-posed one by adding adequate constraints, including joint torque minimization \cite{kawato1990trajectory} and travel distance minimization \cite{tani1996model}.
The current experiment shows an examination of the case using the travel time minimization constraint by adding an additional cost term in the plan free energy $F_{\text{plan}}$ shown previously in Equation~\ref{eq:plan_esfe}.
The modified plan free energy $F'_{\text{plan}}$ is shown in Equation~\ref{eq:ttm}.
\begin{equation} \label{eq:ttm} % (14)
\begin{aligned}
    \gamma_t &= t \log P (s_t|d_t), \\
    F'_e(\bm{x}, \hat{g}, z) &= \sum^{t=t_c}_{t=t_c-win^p} \Big( w \cdot D_{KL}\big[ q(z_t | \bm{x}_{t:t_c}, \hat{g}_{t:t_c}) || p(z_t | d_{t-1}) \big] - E_{q(z_t | \bm{x}_{t:t_c}, \hat{g}_{t:t_c})} \big[ \log P(\bm{x}_t, g_t | d_t) - k\gamma_t \big] \Big), \\
    G'(\hat{g}, z) &= \sum^{t=t_c+win^f}_{t=t_c} \Big( w \cdot D_{KL}\big[ q(z_t | \hat{g}_{t:t_c+win^f}) || p(z_t | d_{t-1}) \big] - E_{q(z_t | \hat{g}_{t:t_c+win^f})} \big[ \log P(\bar{g}_t | d_t) - k\gamma_t \big] \Big), \\
    F'_{\text{plan}} &= F'_e + G'. \\
\end{aligned}
\end{equation}
Where $\gamma$ is the added cost term for minimizing the travel time. 
This cost can be expressed by the summation of the estimated distal probability distribution $P(s_t|d_t)$ multiplied by the time step length at each time step over all time steps in the plan window. 
For this experiment, we set the weight of the travel time cost $k=0.1$.

\begin{figure}[H]
\centering
\includegraphics[width=0.4\linewidth]{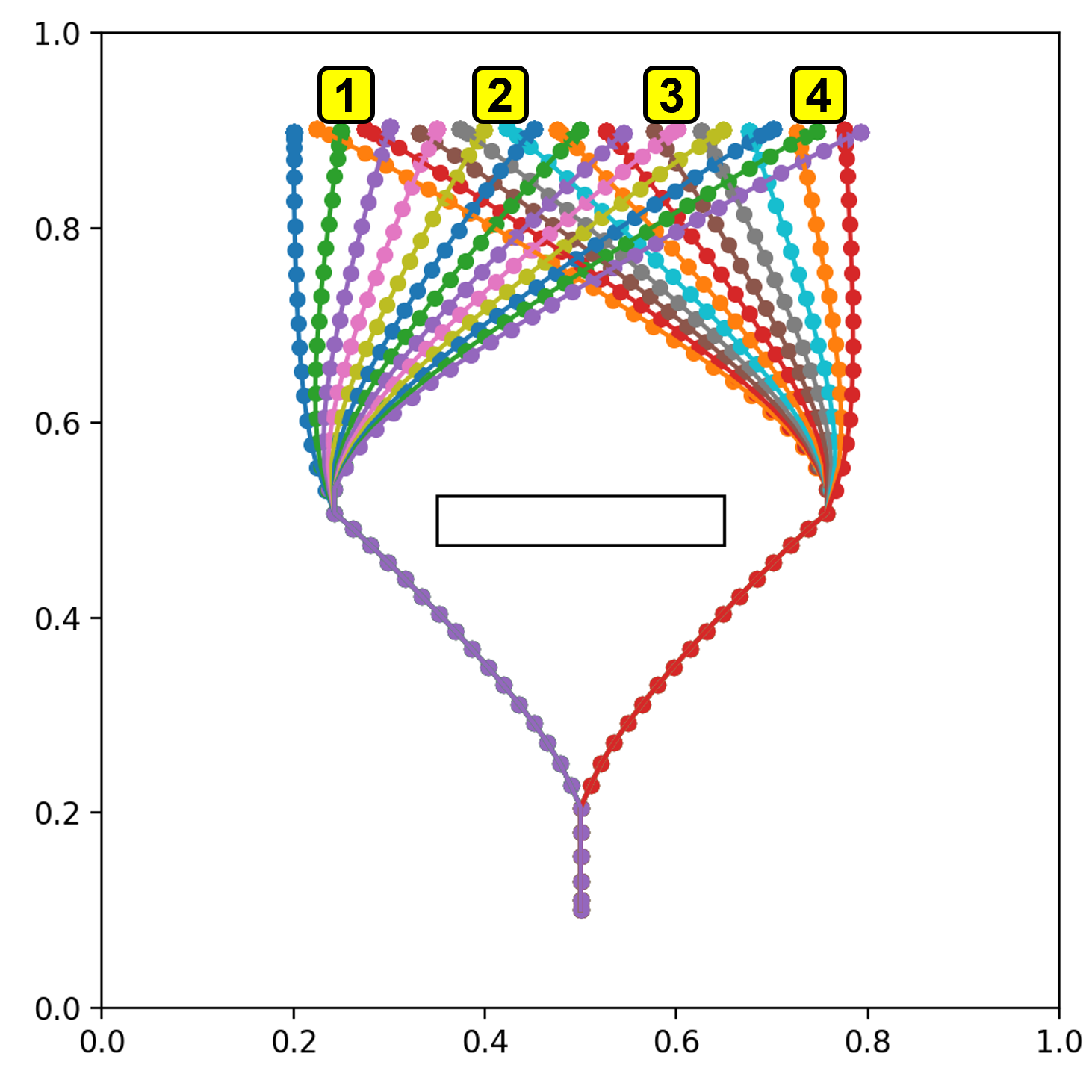}
\caption{Training trajectories overlaid with the four untrained test goals.}
\label{fig:exp1d_data_targets}
\end{figure}

To evaluate the effect of adding the constraint for travel-time minimization, we prepared three separately trained networks and four untrained goal positions that are shown with numbers overlaid on the training trajectories in Figure~\ref{fig:exp1d_data_targets}. The goal positions are selected to avoid the edges (lack of training trajectories) and the center (no difference in trajectory length). 

\begin{figure}[H] % Fig. 17
\centering
\includegraphics[width=0.6\linewidth]{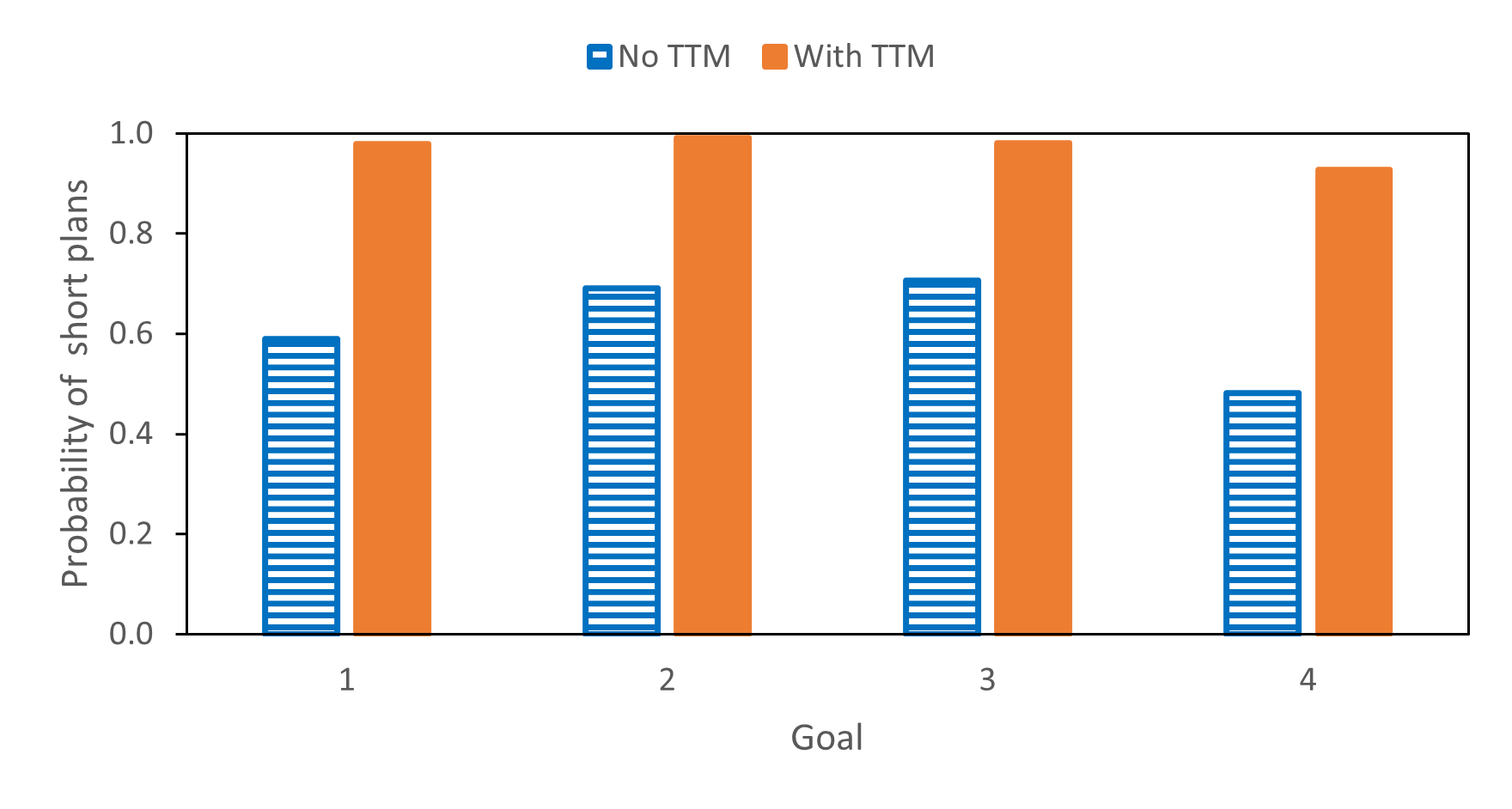}
\caption{Comparison of the probability of shorter plans being generated for each test goal position with and without adding the travel time minimization (TTM).}
\label{fig:exp1c_result}
\end{figure}

The generated action plans for these test goal positions were classified as `short' if the shorter of the possible paths was generated. 
Plan generation was repeated for each test goal position with 1000 different samples.
The resultant probabilities for generating the shorter plans for each goal position with and without travel time cost are shown in Figure~\ref{fig:exp1c_result}. 
Without introducing the travel time cost $\gamma$, the probability of generating the short plans was around 50\%, which is consistent with the ratio in the teaching trajectory set. 
This probability increased to over 90\% with the addition of travel time cost to the plan free energy. These results confirm that the current modified model supports goal-directed plan generation in a well-posed manner.

%%
%% Experiment 2
%%
\subsection{Experiment 2: object manipulation by a physical humanoid robot} \label{sec:exp2}
\begin{figure}[H] % Fig. 18
\centering
\includegraphics[width=0.6\linewidth]{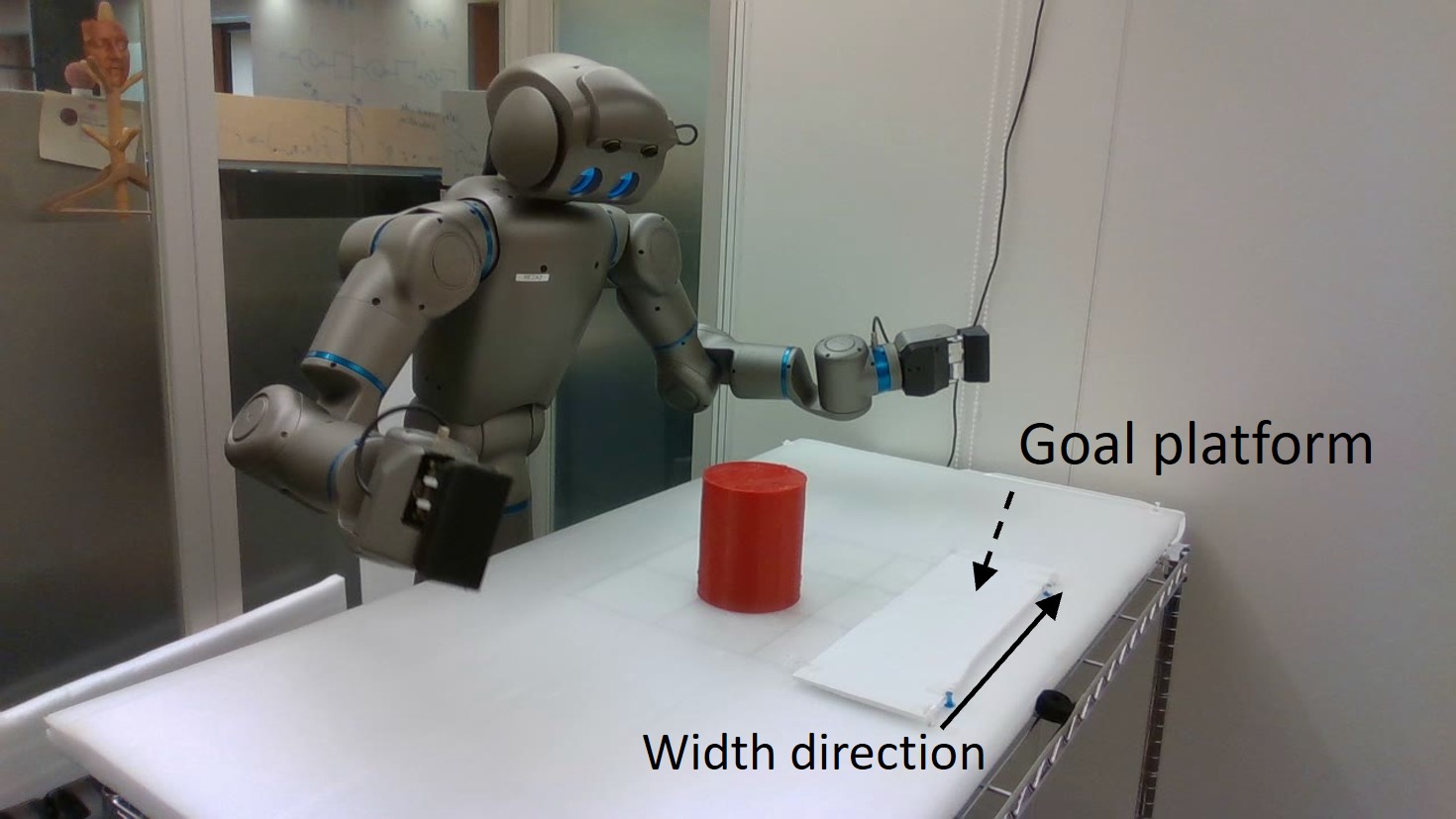}
\caption{The Torobo humanoid robot, with the workspace, goal platform and object.}
\label{fig:exp2_setup}
\end{figure}

In order to verify the performance of the proposed model in a complex physical world, we conducted experiments involving object manipulation by a humanoid robot, Torobo, manufactured by Tokyo Robotics Inc. 
Torobo was placed in front of an object manipulation workspace where a red cylindrical object was located for manipulation.
Two types of goal-directed actions were considered. One was to grasp the object located at an arbitrary position in the workspace (36cm $\times$ 31cm) with both arms and then place it at a specified goal position on the goal platform (42cm wide) fixed at one end of the workspace.
The other type of goal was to grasp the object located at an arbitrary position in the workspace and to swing it up and down.
The Torobo humanoid robot, red cylinder object, workspace, and goal platform are shown in Figure~\ref{fig:exp2_setup}.
The neural network controlled Torobo's two arms (6 degrees of freedom for each arm) and hip joints (2 degrees of freedom) to perform these goal-directed actions (total of 14 joint angles).
The reading of these joint angles represents the proprioception of Torobo.
Torobo can track the position of the red cylinder located in the workspace using a camera mounted in its head.
The object position is constantly tracked by controlling the pitch and yaw of the camera head to keep the target object centered in the camera's field of view. The red cylinder is visually located using YOLOv3 \cite{yolov3}.
Therefore, the pitch and yaw of the head indicates the position of the object, and are considered to represent the exteroception of Torobo.
Thus, exteroception can be represented by only a two-dimensional vector instead of a high-dimensional camera image. This simplification in visual image processing was necessary in order to generate goal-directed actions in real-time.

We conducted two experiments with Torobo. In Experiment 2A, we evaluated the performance in generating goal-directed planning and its execution in a similar fashion to Experiment 1B in Section~\ref{sec:exp1b}, and in Experiment 2B we evaluated the capability of the network for goal inference by observation.
The parameters used for each layer of the RNN are as shown in Table~\ref{tbl:exp2rnn}. As in Experiment 1, the network was trained for 100,000 epochs, using the Adam optimizer with a learning rate $\alpha = 0.001$, $\beta_1=0.9$, $\beta_2=0.999$. 
In order to maintain real-time operation with the robot, planning uses different parameters of $\alpha = 0.1$ and 100 error regression iterations per sensorimotor time step.
\begin{table}[H] % Table 4
\centering
\caption{PV-RNN parameters for Experiment~2. Parameter settings are identical to Experiment~1, only using a larger $\tau$ to compensate for longer sequences.} \label{tbl:exp2rnn}
\begin{tabular}{cccc}
\toprule
                          & \multicolumn{3}{c}{\textbf{Layer}} \\
	                      & \textbf{1}	& \textbf{2} & \textbf{3} \\
\midrule
$\mathbb{R}^d$     	      & 60			& 40         & 20 \\
$\mathbb{R}^z$     	      & 6			& 4          & 2  \\
$\tau$                    & 2           & 10         & 20 \\
$w$                       & 0.0001      & 0.0005     & 0.001 \\
$w_{t=1}$                 & 1.0         & 1.0        & 1.0 \\
\bottomrule
\end{tabular}
\end{table}
The training dataset consists of 120 trajectories for one of the goal-directed actions, grasping then placing, and 80 trajectories for the other type of goal-directed action, grasping then swinging.
The object is located at a random position within the workspace for each sample of the teaching trajectories.
The goal position for placing is also randomly selected along the goal platform's width.
At each sensorimotor time step, we recorded 12 joint angles for both arms and 2 joint angles for the hip joints representing the proprioception and 2 head joint angles representing the exteroception along with a 3D vector representing the preferred goal and 1D scalar for the distal step marker. 
The preferred goal $(\hat{g}^\alpha, \hat{g}^\beta)$ is represented in a similar manner to Experiment 1 wherein $\hat{g}^\beta$ is a 2D one-hot vector representing either goal of grasping-placing or grasping-swinging, and $\hat{g}^\alpha$ is a scalar representing the preferred goal position in the width direction of the goal platform in the case of the grasping-placing goal.

% Experiment 2A
\subsubsection{Experiment 2A: Goal-directed plan generation and execution} \label{sec:exp2a}
\begin{figure}[H]
\centering
\subcaptionbox{\label{fig:exp2a_results_gr}}{\includegraphics[width=0.4\linewidth]{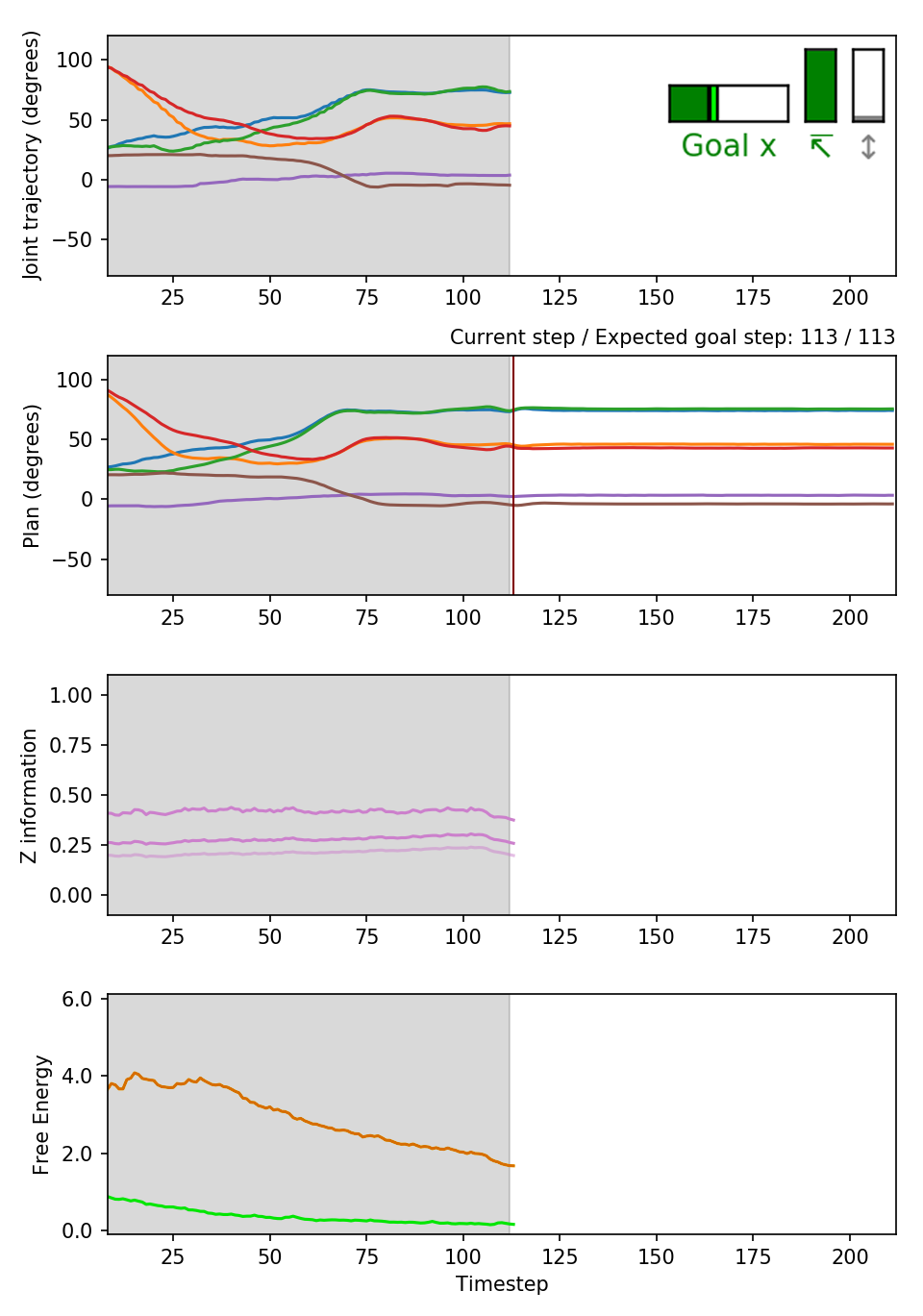}}
\subcaptionbox{\label{fig:exp2a_results_lc}}{\includegraphics[width=0.4\linewidth]{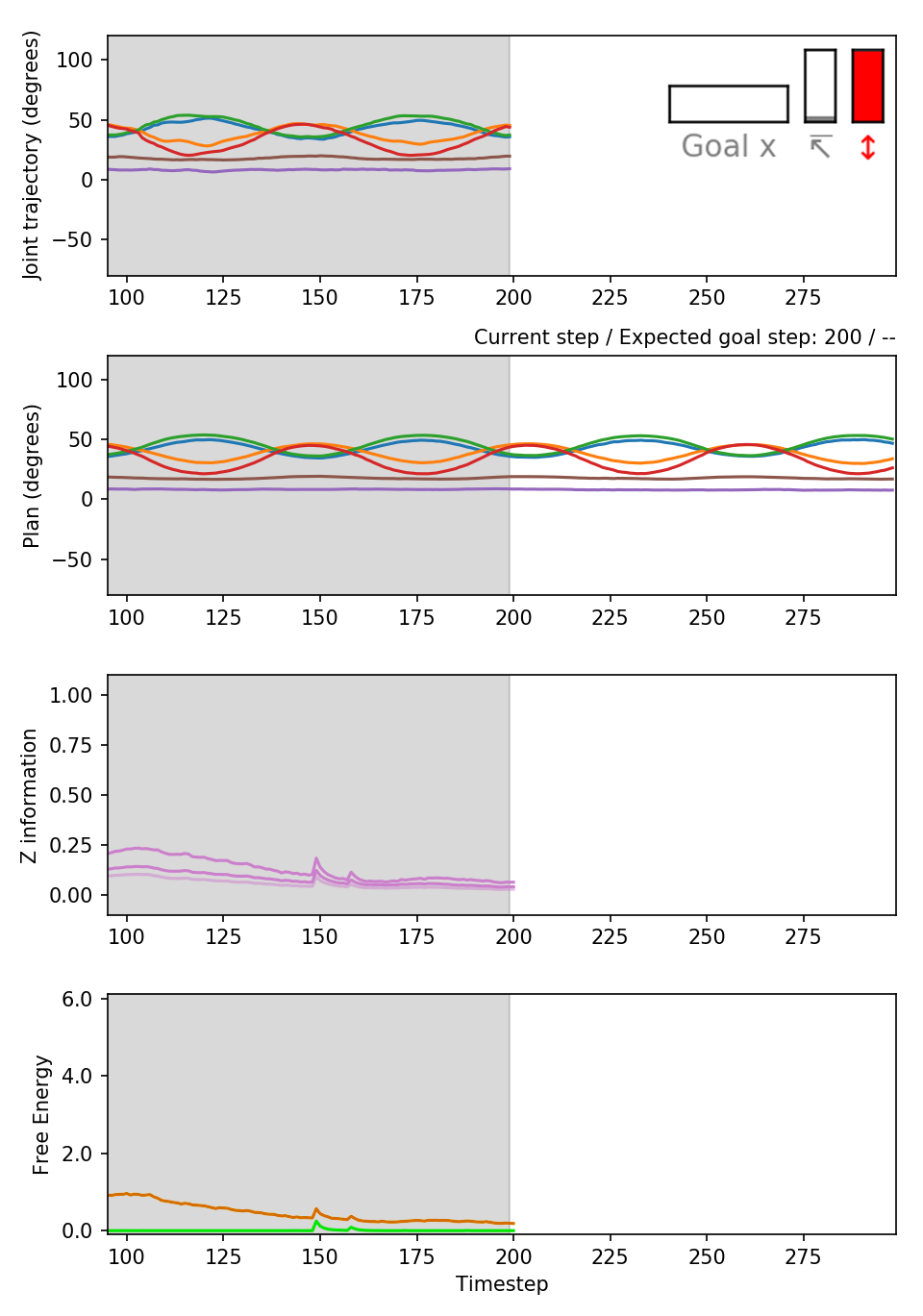}}
\caption{Plots generated while the robot is executing the goal-directed plan. Top to bottom: the sensorimotor history (goal inset on the top right, enlarged for visibility), current generated plan, Z information, and free energy (dark orange: evidence FE, light green: expected FE). Note that only a selected number of joint angles are shown for clarity. (\textbf{a}) An example of grasping-placing, as it reaches the distal step (red line), and (\textbf{b}) an example of grasping-swinging, which can cycle indefinitely.}
\label{fig:exp2a_results_pgen}
\end{figure}

To evaluate the performance of goal-directed plan generation and execution with the physical robot, we measured the RMS deviation to the ground truth, normalized to the data range, as shown in Experiment 1B.
To examine the performance for achieving the grasping and placing goal, the experimenter placed the object at an arbitrary goal position on the goal platform, allowing Torobo to recognize the goal position by visual tracking. 
This position was marked as the ground truth. 
The object was then placed at a random position in the workspace, and the network started to generate action plans to achieve this specified goal while Torobo executed the generated motor plan simultaneously, in real-time.
The difference between the final position of the placed object and the ground truth was then measured for 10 random goal positions. 
In the case of examining the grasping and swinging goal, the object was placed in three different positions in the workspace, and then the resulting robot trajectory in the swinging phase was compared to the closest teaching trajectory in the swinging phase.
The result is shown in Table~\ref{tbl:exp2a_results_pgen}.
\begin{table}[H] %Table 4
\centering
\caption{Deviation from the ground truth for Experiment 2A, given as normalized RMS.} \label{tbl:exp2a_results_pgen}
\begin{tabular}{ccc}
\toprule
	              & \textbf{Grasping-placing} & \textbf{Grasping-swinging} \\
\midrule
NRMSD     	      & 0.10053 		  & 0.01514 \\
\bottomrule
\end{tabular}
\end{table}
Compared to the results obtained in Section~\ref{sec:exp1b}, the network generated a similar low deviation for achieving the grasping and swinging goal, while the deviation was higher for the grasping and placing goal. 
This is likely due to the relatively low precision in tracking the object, especially when placing the object on the goal platform, which was located at the far edge of the workspace.
Figure~\ref{fig:exp2a_results_pgen} presents two plots showing an example of the plans generated when given the two types of goals. Example videos showing these experimental results can be seen at this \href{https://youtu.be/SF8cC68fupA}{link}.

%%%%

% Experiment 2B
\subsubsection{Experiment 2B: Goal understanding} \label{sec:exp2b}
Finally, the experimental results for goal understanding are briefly described.
In this experiment, the object was moved by the experimenter emulating the manipulation of the object by Torobo for each goal type. 
Torobo observed this object movement using object tracking while Torobo remained in the initial posture, except for its head joints, used for object tracking.
The network in Torobo inferred the expected goal and predicted the future movement trajectory in terms of the sensory sequence, as shown in Experiment 1B.
This experiment was repeated 5 times for each of the goal-directed grasping actions, placing and swinging. The network was judged to have correctly inferred the goal if the anticipated goal stably matched the experimenter's actions, before plan execution or the experimenter's actions ended. A video showing this experiment can be seen at this \href{https://youtu.be/iZt0OLQ4I1c}{link}.

The result of this experiment was that the network inferred the goal correctly with 60\% probability while observing the placing actions with 100\% probability while observing the swinging actions. 
However, we note that the capability of the network to correctly infer the goal and future actions requires the actions of the human grasping the cylindrical object to closely match the robot's own learned movement image, which is not easy for humans to consistently reproduce. 
Particularly in the case of grasping-placing, precise timing and position in grasping and placing become critical and the loss of precision in the visual tracking of the object at longer distances posed additional challenges.
When the demonstrated actions deviated from those learned by the network, we observed that the free energy increased instead of decreasing over time, which produced unreliable results.
A future study should investigate methods of making the network more robust in order to better tolerate unreliable human factors, which would be encountered in real world settings.

%%%%%%%%%%%%%%%%%%%%%%%%%%%%%%%%%%%%%%%%%%
\section{Discussion} \label{sec:discussion}
The current study proposed a novel model for goal-directed action, planning, and execution under a teleological framework using the free energy principle.
The proposed model, T-GLean, is characterized by three features. First, goals can be specified either by a specific sensory state expected at a distal step or dynamically changing sensory sequences. Second, goals can be inferred by observed sensory sequences. Third, the goal-directed plan is generated by situating the latent state to the observed sensation by means of the online inference.

The proposed model was evaluated by conducting two experiments, the first using a simulated mobile agent for navigation and the second using a physical humanoid for object manipulation.
The results of experiments using a simulated mobile agent showed that generalization in generation of reaching movements to unlearned goal positions is sufficient with a relatively small number of training samples, with modest improvement as the number of teaching trajectories is increased. It was also shown that both types of goal-directed plan generation and their execution, i.e., reaching a specified position and cycling could be performed precisely. Furthermore, it was demonstrated that goals could be inferred adequately from the observed sensation, even in the case of dynamically changing goals.
Finally, it was shown that the network could generate goal-directed reaching plans with the shortest path when an additional cost for travel time minimization was added to the original plan free energy formula.
This confirms that the current model using this modified plan free energy can generate optimal goal-directed plans under well-posed conditions.

In the results of the experiments scaled up to using a real humanoid robot, it was shown that goal-directed plan generation and execution, as well as goal inference by observation could be performed with reasonable performance for two different goal-directed actions, grasping-placing and grasping-swinging, although their performance was slightly worse compared to the simulated mobile agent case. This could be due to various real world constraints, including limited precision in the visual tracking system and in motor control, as well as unreliable human behavioral factors in demonstrating emulated goal-directed actions to the robot.
There is still plenty of room for improving performance in such real-world situations by making technical efforts in various regards.

Various research topics can be considered for extending the current study in the future.
One interesting topic would be to examine how robots can deal with unexpected environmental changes, using the current model. For example, if Torobo fails to grasp the object or drops it, can it recover from the failure by generating a new recovery action plan? It would be interesting to examine how much such an unexpected situation can be recognized by inferring the latent state by means of minimization of the evidence free energy applied to the past window.
This may require additional learning of various failure situations so that novel situations can be adequately handled through generalization, while maintaining well-posed solutions for normal situations.

Another direction for future study would be further scaling up in action and goal complexity by introducing a language modality.
By using the power of rich linguistic expressions, it is expected that various complex goals can be represented in a compositional manner.
It is, however, very likely that it will be quite difficult to learn an adequate amount of language relevant to goal-directed actions with different levels of complexity at once.
With regard to this problem, one plausible but challenging approach may be the introduction of developmental pathways in learning.
It would be natural to start by learning a set of simple goal representations that could be achieved by some primitive behaviors.
When learning proceeds further, more complex goal-directed actions could be learned by means of compositions of the prior-learned action primitives associated with corresponding compositional linguistic expressions.
This might lead to acquisition of a more abstract goal representation at the conceptual level.

%%%%%%%%%%%%%%%%%%%%%%%%%%%%%%%%%%%%%%%%%%
\vspace{6pt} 

\bibliographystyle{unsrtnat}
\bibliography{ref}

\end{document}